\documentclass[letterpaper]{article} 
\usepackage{aaai24}  
\usepackage{times}  
\usepackage{helvet}  
\usepackage{courier}  
\usepackage[hyphens]{url}  
\usepackage{graphicx} 
\usepackage[numbers]{natbib}  
\usepackage{subcaption}
\usepackage{newfloat}
\usepackage{listings}
\DeclareCaptionStyle{ruled}{labelfont=normalfont,labelsep=colon,strut=off} 
\nocopyright 

\usepackage{amssymb}
\usepackage{amsmath}
\usepackage{dsfont}
\usepackage{nicefrac}
\usepackage{booktabs} 
\usepackage{threeparttable}
\usepackage{pifont}
\usepackage{multirow}

\usepackage[dvipsnames,table]{xcolor}

\newcommand{\LS}[1]{\left({#1}\right)}
\newcommand{\LM}[1]{\left\{{#1}\right\}}
\newcommand{\LL}[1]{\left[{#1}\right]}
\newcommand{\Mid}{\;\middle|\;}
\newcommand{\C}[1]{\mathcal{#1}}
\newcommand{\R}{\mathbb{R}}
\newcommand{\BB}[1]{\mathbb{#1}}
\newcommand{\B}[1]{\mathbf{#1}}

\newcommand{\round}[1]{\left\lfloor #1\right\rceil}
\newcommand{\normx}[1]{\left\Vert#1\right\Vert}

\usepackage[ruled,vlined,linesnumbered]{algorithm2e}
\makeatletter
\newcommand{\algorithmfootnote}[2][\footnotesize]{%
  \let\old@algocf@finish\@algocf@finish
  \def\@algocf@finish{\old@algocf@finish
    \leavevmode\rlap{\begin{minipage}{\linewidth}
    #1#2
    \end{minipage}}%
  }%
}

\newcommand{\app}{\raise.17ex\hbox{$\scriptstyle\sim$}}

\definecolor{codegreen}{rgb}{0.0,0.6,0.0}
\definecolor{lightgreen}{rgb}{0.62, 1.00, 0.84}

\title{SIRA: Scalable Inter-frame Relation and Association for Radar Perception}
\author{
    Ryoma Yataka\textsuperscript{\rm 1,2},  
    Pu (Perry) Wang\textsuperscript{\rm 1}, 
    Petros Boufounos\textsuperscript{\rm 1}, 
    Ryuhei Takahashi\textsuperscript{\rm 2}
}
\affiliations{
    \textsuperscript{\rm 1}Mitsubishi Electric Research Laboratories (MERL), Cambridge, MA 02139, USA\\
    \textsuperscript{\rm 2}Information Technology R\&D Center, Mitsubishi Electric Corporation, Kanagawa 247-8501, Japan
    {\tt\small  \{yataka,pwang,petrosb\}@merl.com, Takahashi.Ryuhei@ab.MitsubishiElectric.co.jp}
}

\begin{document}

\maketitle

\begin{abstract}
Conventional radar feature extraction faces limitations due to low spatial resolution, noise, multipath reflection, the presence of ghost targets, and motion blur. Such limitations can be exacerbated by nonlinear object motion, particularly from an ego-centric viewpoint. It becomes evident that to address these challenges, the key lies in exploiting temporal feature relation over an extended horizon and enforcing spatial motion consistency for effective association. To this end, this paper proposes SIRA (Scalable Inter-frame Relation and Association) with two designs. First, inspired by Swin Transformer, we introduce extended temporal relation, generalizing the existing temporal relation layer from two consecutive frames to multiple inter-frames with temporally regrouped window attention for scalability. Second, we propose motion consistency track with the concept of a pseudo-tracklet generated from observational data for better trajectory prediction and subsequent object association.  Our approach achieves $58.11$ mAP$@0.5$ for oriented object detection and $47.79$ MOTA for multiple object tracking on the \textit{Radiate} dataset, surpassing previous state-of-the-art by a margin of $+4.11$ mAP$@0.5$ and $+9.94$ MOTA, respectively. 
\end{abstract}

\section{Introduction}
\label{sec:intro}

Automotive perception involves the interpretation of the external driving environment and internal vehicle cabin conditions with an array of perception sensors to achieve robust safety and driving autonomy~\cite{Pandharipande2023_SensingMachineLearning}. Compared to optical camera and lidar sensors, radar is cost-effective, friendly to sensor maintenance and calibration, and has distinct advantages in providing long-range perception capabilities in adverse weather and lighting conditions~\cite{Zeng2014_AutomotiveRadarWeather}.  

Nevertheless, a notable limitation of radar-based automotive perception is its low spatial resolution in the azimuth and elevation domains, and its inherent noise including multipath reflection, ghost targets, and motion blur. As a result, its ability to detect and track objects lags behind the requirements for fully autonomous driving capabilities. Recently, standalone radar-only perception has been investigated in~\cite{Zhang2021_RADDet, Ouaknine2021_RadSemSeg, Gao2021_RAMP_CNN, Bai2021_RadarTransformer, Li2022_TemporalRelations, Palffy2022_PointPillar, Li2023_CVPR_RADDet}. Li et al.~\cite{Li2022_TemporalRelations} proposed a framework called TempoRadar to study temporal attention to features from $2$ ego-centric bird-eye-view (BEV) radar frames. It has shown promising performance gains when evaluated on the large-scale open \textit{Radiate}~\cite{Sheeny2021_RADIATE} dataset.

\begin{figure}[t]
    \centering
    \includegraphics[width=0.46\textwidth]{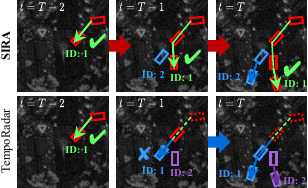}
    \caption{
        Conventional radar perception pipelines such as TempoRadar \cite{Li2022_TemporalRelations} (Bottom Row) rely on a limited number (one or two) of frames and the limited time horizon may lead to incorrect feature-level and object-level association (e.g., $t=T-1$) and propagate to subsequent frames (e.g., $t=T$). In contrast, SIRA (Top Row) accounts for joint spatio-temporal consistency over an extended temporal horizon (e.g., all $3$ frames here), allowing for more accurate association in nonlinear motion scenarios even in an ego-centric viewpoint.   
    }
    \label{fig:concept}
    \vspace{-5mm}
\end{figure}
However, such limitations can be exacerbated by nonlinear object motion, particularly from an ego-centric BEV.
In particular, low frame rates result in significant influence from the nonlinearity of object motion, leading to frequent tracking errors. Conventional radar perception pipelines such as TempoRadar enables prediction based on information from the previous frame, but in the case of objects with fast and nonlinear motion within radar frames, such information is inadequate (Bottom of Fig.~\ref{fig:concept}). Although applying Kalman filter (KF~\cite{Klmn1960_KF})-based algorithms~\cite{Bewley2016_SORT,Zhang2022_ByteTrack,Cao2023_OCSORT,Du2023_StrongSORT}, is possible, radar perception is difficult to relate accurately due to a complex combination of factors, including the effects of high-speed nonlinear motion dynamics and the lack of detailed appearance features due to low resolution. To address these limitations and improve radar perception for object detection and tracking, we propose a framework called \textit{\textbf{scalable inter-frame relation \& association (SIRA)}}. SIRA consists of two modules: extended temporal relation (ETR) and motion consistency track (MCTrack).
The contributions of this study are as follows:
\begin{itemize}
    \item We introduce ETR, generalizing the existing temporal relation layer from two consecutive frames to multiple inter-frames with temporally regrouped window attention for scalability. It emphasizes the temporal consistency of moving objects by enabling accurate detection while maintaining computational efficiency over long time horizon. This can facilitate easy detection through consistent correlations across multiple frames at the object level.
    \item We designed MCTrack based on the concept of pseudo-tracklets, which are generated by using a learnable module to predict the arbitral nonlinear motion of an object between multiple frames, and the association caused by these pseudo-tracklets enhances spatial consistency during inference. Thus, MCTrack enables more reliable position predictions, even in scenarios with fast-moving objects and low frame rates.
    \item We propose \textit{\textbf{SIRA}} that adopts a loss function for the end-to-end learning of these two modules, achieving stable predictions that capture the spatio-temporal consistency of nonlinear moving objects.
    \item We evaluate SIRA on \textit{Radiate}~\cite{Sheeny2021_RADIATE}, a BEV radar dataset. Our approach achieves $58.11$ mAP$@0.5$ for oriented object detection and $47.79$ MOTA for multiple object tracking on the \textit{Radiate} dataset, surpassing previous state-of-the-art by a margin of $+4.11$ mAP$@0.5$ and $+9.94$ MOTA, respectively. 
\end{itemize}

\section{Related Work for Radar Perception}
\label{sec:related_works}
Automotive radar predominantly employs a frequency-modulated continuous waveform (FMCW) for object detection, generating point clouds. The fundamental of FMCW is explained in Appendix~\ref{sec:fundamentals_FMCW}.
In addition, we defer a short review of recent visual tracking in Appendix~\ref{sec:visual_tracking}.

\paragraph{Detection by Radar:}
For automotive perception, radar-assisted multimodal approaches were proposed~\cite{Teck2019_RadarCameraFusionForVehDet, Yang2020_RadarNet, Qian2021_LidarRadarRobustMultimodal, Wang2023_CVPR_BiLRFusion, Ding2023_CVPR_HiddenGems,Man2023_BEVGuidedMultiModalFusion}.
Compared with multimodals, standalone radar-only perception has been studied in~\cite{Zhang2021_RADDet, Ouaknine2021_RadSemSeg, Gao2021_RAMP_CNN, Bai2021_RadarTransformer, Li2022_TemporalRelations, Palffy2022_PointPillar, Li2023_CVPR_RADDet,Fent2023_RadarGNN}. 
A multi-view feature fusion method was proposed in~\cite{Gao2021_RAMP_CNN} to combine features from range-Doppler, range-angle, and angle-Doppler radar heatmaps for object classification. As opposed to single-frame radar feature extraction, Li et al.~\cite{Li2022_TemporalRelations} proposed TempoRadar with 2 frames. 

\paragraph{Mutiple Object Tracking by Radar:}
Object tracking with radar has seen several proposals depending on the sparsity or density of the radar points obtained for each object~\cite{Pandharipande2023_SensingMachineLearning}. For sparse radar detection points, model-based tracking algorithms have been explored in the context of extended object tracking (EOT)~\cite{Karl2016_BEOT0}. They use Bayesian filtering~\cite{Koch2008_BEOT1,Orguner2012_BEOT2,Baum2014_BEOT3,Wahlstrom2015_BEOT4,Bro2017_BEOT5,Karl2020_BEOT6,Xia2021_BEOT7} to model the spatial distribution of radar detection points across the vehicle's range and predict and update the extended states such as position and velocity. 
Moreover, to address the nonlinearity problem due to objects deviating from constant linear motion, algorithms such as extended KF~\cite{Smith1962_EKF} and unscented KF~\cite{Julier1997_UKF} have been proposed to handle nonlinear motion using first- and third-order Taylor approximations. However, these still rely on approximating the Gaussian prior distribution assumed by the KF, making modeling challenging for movements where the next position is determined by human intent, such as in vehicles. Particle filter~\cite{Gustafsson2002_PF} addresses nonlinear motion using a sampling-based posterior estimation, which requires exponential computation.
For high-density radar detection points, following~\cite{Yin2021_Center3DTracking,Zhou2020_CenterTrack}, TempoRadar extended the achieved strong tracking performance through learning. 
Our proposed framework extends KF-based methods and learning-based approaches by assuming high-density radar detection points. It explicitly considers strong object-level consistency by using multiple frames to capture the nonlinear motion of objects.

\begin{figure*}[t]
    \centering
    \includegraphics[width=\textwidth]{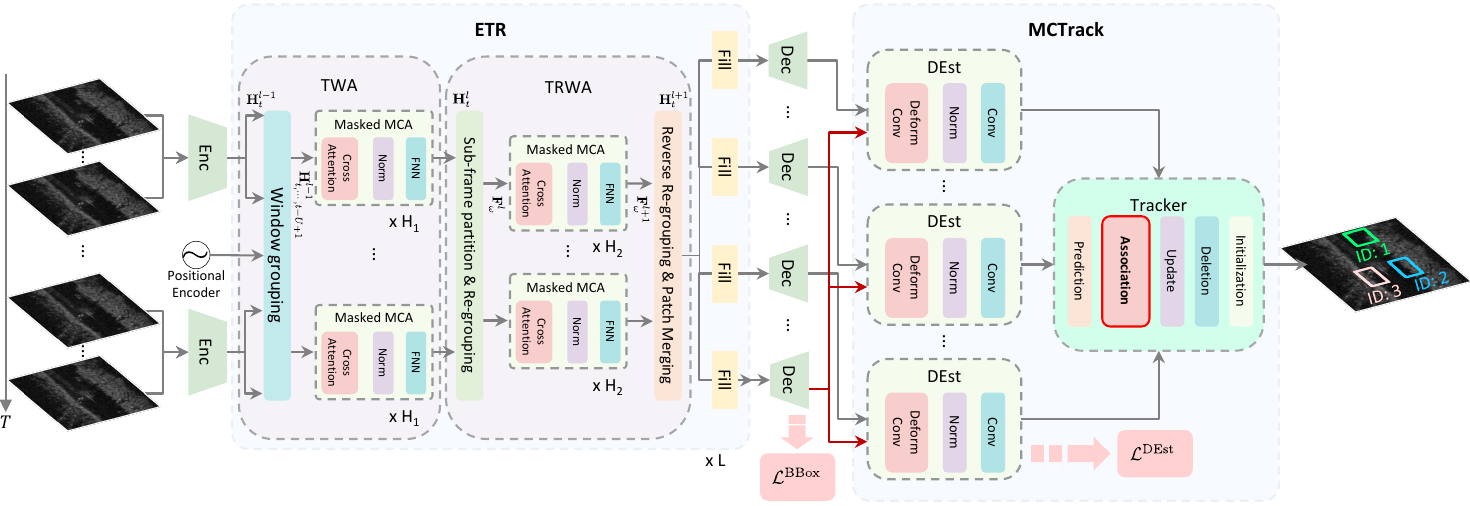}
    \caption{
        The architecture of SIRA with two modules:~1) extended temporal relation (ETR) capturing the temporal feature consistency while maintaining computational efficiency,  and 2) motion consistency track (MCTrack) estimating pseudo-direction of objects during training and establishing pseudo-tracklets for better association in inference. The detection loss $\C{L}_{t}^{\text{BBox}}$ and pseudo-direction loss $\C{L}^{\text{DEst}}$ are used to train the pipeline end-to-end for object detection and tracking. 
    }
    \label{fig:concept_architecture}
    \vspace{-1mm}
\end{figure*}

\section{Scalable Inter-frame Relation \& Association}
\label{sec:SIRA}
An overview of the SIRA framework is illustrated in Fig.~\ref{fig:concept_architecture} with two main modules: 1) ETR and 2) MCTrack. ETR focuses on the temporal consistency, while MCTrack captures the spatial motion consistency, ensuring the continuity and accuracy of object detection and tracking at the output.  

\subsection{Preliminary}
\label{sec:basic_framework}

\paragraph{Encoder:}
Radar perception pipelines employ an encoder to transform the radar frame $I_t \in \R^{1\times H\times W}$ into high-level features and accentuate the position of objects.
\begin{equation}
    \B{Z}_{t} := \C{F}_{\theta}\LS{I_{t}}\in\R^{C\times\frac{H}{s}\times\frac{W}{s}},
\end{equation}
where $C$, $H$, $W$, and $s$ represent the number of channels, height, width, and downsampling ratio over the spatial dimension, respectively. $\C{F}_{\theta}\LS{\cdot}$ is encoder such as ResNet~\cite{He2016_ResNet} with parameters $\theta$.
By denoting multiple $T$ radar frames as $\B{I} = \LM{I_t}_{t=1}^{T} \in \R^{T\times H\times W}$, we can obtain informative features $\B{Z}_{t} =\C{F}_{\theta}\LS{\B{I}}$.

\paragraph{Decoder:}
The decoder estimates the bounding boxes from the features. To localize objects, the two-dimensional (2D) center coordinates $\LS{x_t, y_t}$ of the top-$K$ peak values $\widehat{c}_t$ in the heatmap, corresponding width $\widehat{w}_t$ and length $ \widehat{h}_t$, orientation $\widehat{\vartheta}_t$, and 2D offsets $(\widehat{o}_{x,t}, \widehat{o}_{y,t})$ are predicted as the output bounding box of an object with decoder heads $\C{G}_{\theta}$ as:
\begin{equation}
    \left(x_t, y_t, \widehat{w}_t, \widehat{h}_t, \widehat{\vartheta}_t, \widehat{o}_{x,t}, \widehat{o}_{y,t}, \widehat{c}_t \right)^{\top} = \C{G}_{\theta}\LS{\B{Z}_{t}}.
\end{equation}
One such decoder is the one used in CenterPoint~\cite{Xingyi2019_CenterPoint}. 

\paragraph{Exploiting Temporality:}
For radar perception, it is necessary enhance the feature extraction utilizing additional properties from the temporal domain.  One straightforward way is to stack multiple frames as the input to the encoder, i.e., $\B{Z}_{t} =\C{F}_{\theta}\LS{\B{I}}$. To exploiting the feature-level temporal relation, TempoRadar~\cite{Li2022_TemporalRelations} introduces a temporal relation layer (TRL) that selects top-$K$ features $\B{H}_t \in \R^{C \times K} $ from $\B{Z}_{t} := \C{F}_{\theta}\LS{{I_{{t},{t-1}}}}$ and $\B{H}_{t-1}\in \R^{C \times K} $ from $\B{Z}_{t-1} := \C{F}_{\theta}\LS{{I_{{t-1},{t}}}}$, where ${I_{{t-1},{t}}}$ concatenates two consecutive radar frames along the channel dimension in the order of $\LS{t-1, t}$ with the following feature selector $\C{S}_K$:
\begin{align}
    \label{eq:topK_selector}
    \B{H}_t = \C{S}_K\LS{\B{Z}_t}, \quad t=\LM{t-1, t}.
\end{align}
By concatenating the $2K$ selected features as $\B{H}_{t,t-1}=[\B{H}_{t}, \B{H}_{t-1}]^{\top}$, TRL further computes masked multi-head cross-attention (MCA) as 
\begin{align} \label{eq:attention_of_ETR}
     \C{A}\LS{\B{V}, \B{X}} := \operatorname{softmax}\LS{\frac{\B{M} +  q\LS{\B{X}} k\LS{\B{X}}^{\top}}{\sqrt{d}}} v\LS{\B{V}}
\end{align}
where $\B{V}=\B{H}_{t,t-1}$, $\B{X}=\B{H}^{\text{pos}}_{t,t-1}$ is the concatenated feature $\B{H}_{t,t-1}$ supplemented by the positional encoding, $\{ q\LS{\cdot}, k\LS{\cdot}, v\LS{\cdot}\}$ are query/keys/values, and $d$ is the query/key dimension. The masking matrix $\B{M}$ is designed to turn off the attention between features from the same frame and allow for only cross-frame feature attention to ensure temporal feature consistency. 

These enhanced features are refilled back to $\B{Z}_{t}$ and $\B{Z}_{t-1}$ at corresponding spatial coordinates and fed to the decoder for object detection and tracking.  Refer to Appendix~\ref{sec:temporadar} for the top-$K$ feature selector $\C{S}_K$ and the design of $\B{M}$.  

\subsection{ETR: Extended Temporal Relation}
\label{sec:ETR}
The ETR module borrows the concept of shifted window attention in Swin Transformer \cite{Liu2021_SwinTransformer} but in a deformable temporal fashion. It generalizes the TRL over a longer time horizon of consecutive frames with a scalable complexity. In the following, we introduce the two main blocks: temporal window attention (TWA) and temporally regrouped window attention (TRWA) of ETR shown in Fig.~\ref{fig:concept_architecture}. 

\begin{figure}[t]
    \centering
    \includegraphics[width=0.47\textwidth]{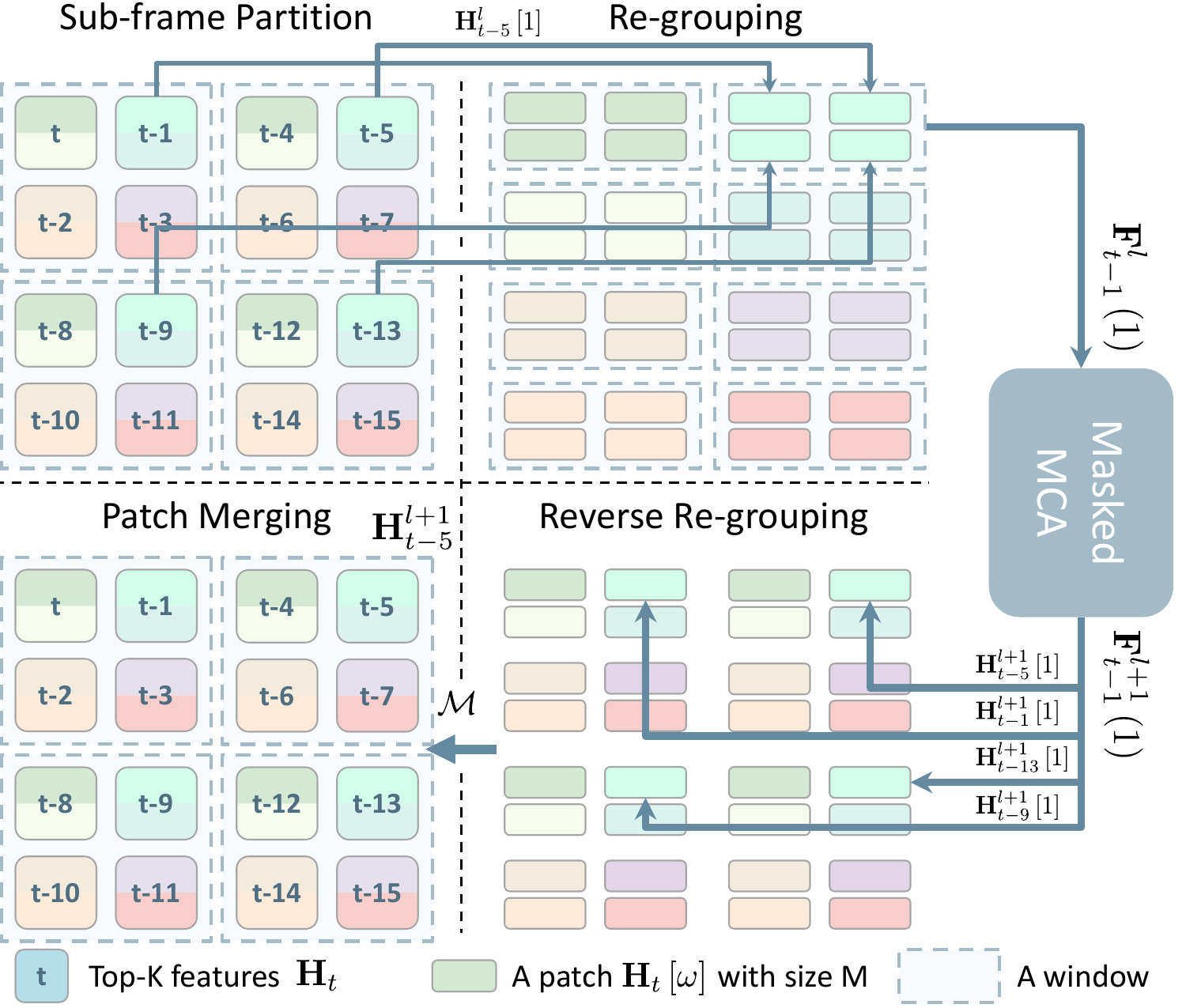}
    \caption{
        The TRWA block of the ETR module. Each frame is partitioned into sub-frame patches (in two contrasting colors of each frame in Top Left) and these patches are regrouped into new windows (Top Right) in a deformable temporal order (arrow lines). Masked multi-head cross-attention (MCA) is applied to new regrouped windows for scalable cross-window attention. 
    }
    \label{fig:ETR_architecture_block2}
    \vspace{-4mm}
\end{figure}

\paragraph{Temporal Window Attention:}
The $l$-th TWA layer expands the TRL from $2$ consecutive frames to a temporal window of $U \geq 2$ frames and computes masked MCA within each window. In Fig.~\ref{fig:ETR_architecture_block2}, we group $U=4$ consecutive frames into one temporal window (in dash boxes) and we have $4$ windows for $T=16$ frames. 

For each temporal window $\{t, t-1, \cdots, t-U+1\}$, we cyclically shift the frame indices and concatenate the $U$ shifted radar frames along the channel dimension for the backbone feature extraction, i.e.,
\begin{align}
\B{Z}_{t} & := \C{F}_{\theta}\LS{{I_{t, t-1, \cdots, t-U+1}}}, \notag \\
\B{Z}_{t-1} &  := \C{F}_{\theta}\LS{{I_{t-1, t-2, \cdots, t-U+1, t}}}, \cdots, \notag \\
 \B{Z}_{t-U+1} & := \C{F}_{\theta}\LS{{I_{t-U+1, t, t-1, \cdots, t-U+2}}}.  
\end{align}
It is easy to see that, when $U=2$, this reduces to the TRL. 
We then follow the same top-$K$ feature selector as the TempoRadar (refer to Appendix~\ref{sec:temporadar})
\begin{align} \label{H0t}
\B{H}_t = \C{S}_K\LS{\B{Z}_t}, \quad t=\{ t, t-1, \cdots, t-U+1\}.
\end{align}

By concatenating features from the temporal window of $U$ frames,  we have $\B{H}_{{t,\cdots,t-U+1}}^{l-1}= [\B{H}^{l-1}_{t}, \cdots, \B{H}^{l-1}_{t-U+1}]^{\top}$, where the superindex denotes the layer index in the ETR model and $\B{H}^{0}_{t}$ takes $\B{H}_t$ of \eqref{H0t} as input for the first layer. 
We apply the masked MCA of (\ref{eq:attention_of_ETR}) $H_1$ times to $\B{H}_{{t,\cdots,t-U+1}}^{l-1}$ with a masking matrix $\B{M}$ of dim $UK\times UK$ for cross-frame feature attention within each window. Collecting from all windows, the TWA block obtains the features $\B{H}_{t}^{l}, \cdots, \B{H}_{t-T+1}^{l}$ from all $T$ frames at its output.

\paragraph{Temporally Regrouped Window Attention:}
To allow for cross-window attention, we regroup the subset features from different windows in a deformable temporal order. First, we partition the $K$ features of each frame into $\Omega$ sub-frame patches with a stride $S$. Each sub-frame patch consists of $M$ features. As shown in Fig.~\ref{fig:ETR_architecture_block2},  one choice for a non-overlapping sub-frame partition is $M = K/2$ and $S = K/2$ (assuming $K$ is even) where each frame is partitioned into $\Omega=2$ sub-frame patches, as illustrated in two contrasting colors for each frame in Fig.~\ref{fig:ETR_architecture_block2}. Alternatively, we may choose $S<M$ for overlapping partition. The resulting sub-frame patches of frame $t$ are defined as $\B{H}_{t}^{l}\LL{\omega}\in\R^{C\times M}, \omega=1, \cdots, \Omega$.  
For more discussion of patch size, refer to Appendix~\ref{sec:training_inference_SIRA}.

The sub-frame patches are regrouped into a new set of windows in a deformable temporal order for cross-window attention. For the newly regrouped window, the features are aggregated as
\begin{equation}
    \B{F}_{t}^{l}(\omega) := \LM{\B{H}_{t}^{l}\LL{\omega}, \B{H}_{t-U}^{l}\LL{\omega}, \cdots, \B{H}_{t-T+U}^{l}\LL{\omega}}^{\top},    
\end{equation}
As illustrated in the top right portion of Fig.~\ref{fig:ETR_architecture_block2}, the regrouping operation extracts one sub-frame patch from each window and results in $U=4$ patches and $UM=UK/2$ features in each new window. Subsequently, we apply the masked MCAs of \eqref{eq:attention_of_ETR} $H_2$ times over the aggregated feature $\B{F}_{t}^{l}(\omega)$ in each new window with an affordable cross-window attention complexity of $TM/U \times TM/U$. 

The cross-window attentive features are re-grouped in the reverse manner to construct the $K$ features of each frame according to the temporal ($t$) and patch ($\omega$) indices. In the case of overlapping patch partitioning, i.e., $S<M$, a patch merging operation $\C{M}$ is necessary to merge the features $\B{H}_{t}^{l+1}=\C{M}\{\B{H}_{t}^{l+1}[1], \cdots, \B{H}_{t}^{l+1}[\Omega]\}$ at the overlapping positions. Patch merging operations (mean, sum and max) will be examined in Section~\ref{sec:ablation_study}. The TRWA block outputs $\B{H}^{l+1}_{t}, \cdots, \B{H}^{l+1}_{t-T+1}$ for all $T$ frames, sharing the same dimension as the input $\B{H}^{l}_{t}, \cdots, \B{H}^{l}_{t-T+1}$. 

\paragraph{Stacking as a Stage:}
We can stack the TWA and TRWA blocks as one stage and repeat the stage $L$ times. In between stages, the output of TRWA block serves the input to the TWA block in the next stage. Finally, we put these features $\B{H}^{l+1}_{t}, \cdots, \B{H}^{l+1}_{t-T+1}$ back to $\{ \B{Z}_{t}, \cdots, \B{Z}_{t-T+1}\}$ at corresponding spatial coordinates.
The effect of $L$ will be examined in Section~\ref{sec:ablation_study}. 

\paragraph{Complexity Analysis:}
For a given $T$,  $K$, and the number of stages $L$,  the computational complexity expressions for TempoRadar \cite{Li2022_TemporalRelations}
and the ETR module are shown below
\begin{align}
    \label{eq:computational_complexity_TR}
    & \text{TempoRadar:} \LS{T K}^2 L\\
    \label{eq:computational_complexity_ETR}
   & \text{ETR:} \LS{\text{TWA} + \text{TRWA}}L= {K^2TUL + MT^2KL/U} 
\end{align}
where $U$ is the number of frames in one temporal window in the TWA block and $M$ is the number of features for each sub-frame patch in the TRWA block. 
Note that, if $U=T$ and $M=K$, ETR reduces to the TWA module only, resulting in a full-size attention like TempoRadar. In this case, the ETR complexity in \eqref{eq:computational_complexity_ETR} reduces to that of TempoRadar in \eqref{eq:computational_complexity_TR}. 
Appendix~\ref{sec:complexity_analysis} provides numerical comparison of the complexity in several settings. 

\subsection{MCTrack: Motion Consistency Track}
\label{sec:tracking}
As shown in Fig.~\ref{fig:concept_architecture}, MCTrack takes the temporally enhanced features $\{ \B{Z}_{t}\}$ from the ETR output, and applies the decoding heads on each $\B{Z}_{t}$ for bounding box estimation. To further exploit motion consistency, we introduce two MC modules: one for training and one for inference, for improved detection and tracking performance. 

\paragraph{Motion Consistency for Training:}
We introduce the concept of \textbf{pseudo-direction} to improve motion consistency during training. Pseudo-directions are vectors that directly predict the current object position from each of the previous frames, using a decoder head with learnable parameters. 
It is used to iteratively refine object positions between frames during learning and the pseudo-direction loss contributes to the overall training loss in Section~\ref{sec:loss}. 

\begin{figure}[t]
    \centering
    \includegraphics[width=0.46\textwidth]{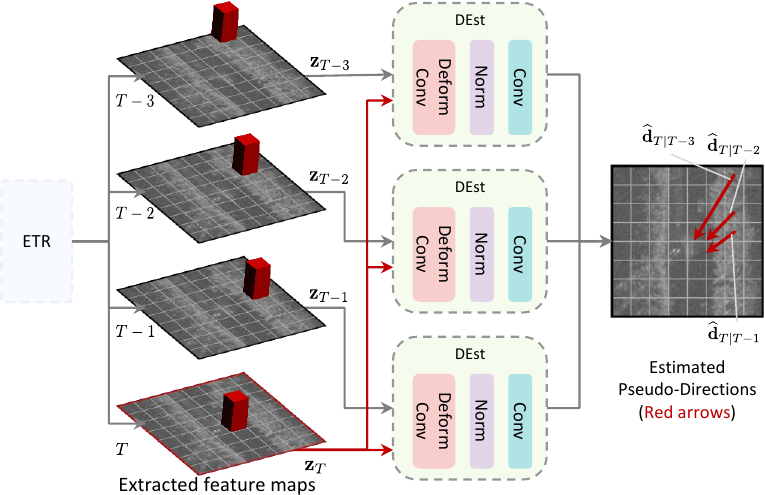}
    \caption{
        Direction Estimation (DEst) decoder head. Each DEst head takes a pair of $2$ frames $\B{Z}_T$ and $\B{Z}_{T-\tau}$, and estimates the pseudo-direction $\widehat{\B{d}}_{T\mid T-\tau}$ (arrow lines in red). 
    }
    \label{fig:tracking_architecture}
    \vspace{-4mm}
\end{figure}

To compute the $\tau$-step pseudo-direction $\widehat{\B{d}}_{T \mid T-\tau}$\footnote{With slightly abused notation,  we use $T$ to denote not only the number of frames, but also current frame index in this section.} from the past frame $T-\tau$ to  frame $T$, we design a specific decoder head $\C{G}_{\theta}^{\text{DEst}}\LS{\cdot}$: direction estimation (DEst) with learnable parameters $\theta$ in Fig.~\ref{fig:tracking_architecture}, 
\begin{equation}
    \widehat{\B{d}}_{T \mid T-\tau} = \C{G}_{\theta}^{\text{DEst}}\LS{\B{Z}_{T},\B{Z}_{T-\tau}}\LL{\B{p}_{\B{z}_T}} \in \R^{2},
\end{equation}
where $\B{Z}_T$ and $\B{Z}_{T-\tau}$ are temporally enhanced features at frame $T$ and $T-\tau$,  $\B{p}_{\B{z}_T}$ is a two-dimensional coordinate, and $\tau=1, 2, \cdots, T-1$. Fig.~\ref{fig:tracking_architecture} shows the DEst head architecture, comprising the deformable convolution~\cite{Dai2017_DeformConv}, normalization, and convolution layers. 
The deformable convolution is particularly used to capture features of objects that have undergone significant displacement across $\tau$ frames.

The estimated vectors represent the positional differences of objects across $\tau$ frames. It is essential to address scenarios where objects move significantly within just one frame due to low frame rates and ego-vehicle motions. 

\paragraph{Motion Consistency for Inference:}

\begin{figure}[t]
    \centering
    \includegraphics[width=0.46\textwidth]{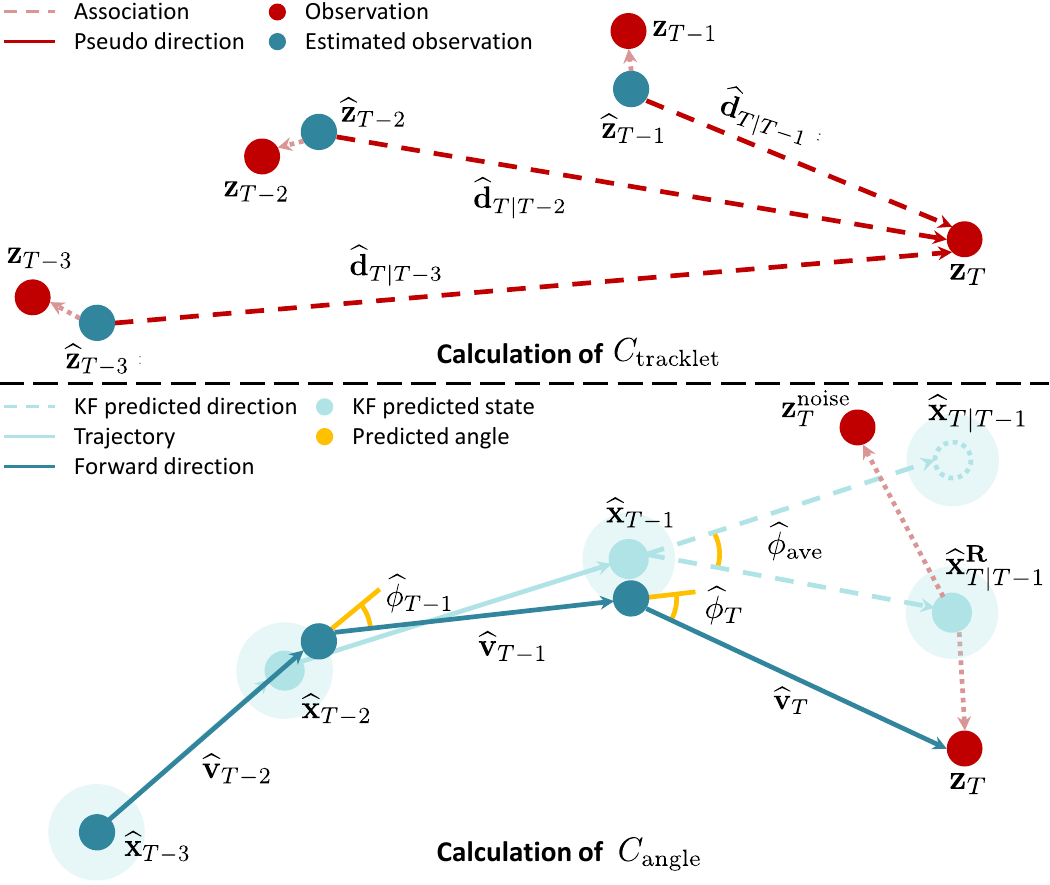}
    \caption{
        The calculation of similarity metrics $C^{\text{angle}}$ and $C^{\text{tracklet}}$ in MCTrack at inference. A pseudo-tracklet $\{\LM{\widehat{\B{z}}_t}_{t=1}^{T}, \LM{\widehat{\B{v}}_t}_{t=2}^{T}\}$ is constructed with $\widehat{\B{d}}_{T|T-\tau}$ estimated with DEst, and is used for association: (Top) rotating a state $\B{x}_{T|T-1}$ to be more correlate the observation $\B{z}_T$, (Bottom) directly correlating the observations $\B{z}_t$ with $\widehat{\B{z}}_t$.
    }
    \label{fig:concept_association}
    \vspace{-3mm}
\end{figure}

In inference, we use a KF-based tracker such as OC-SORT~\cite{Cao2023_OCSORT} to enforce motion consistency. As shown in Fig.~\ref{fig:concept_architecture}, the tracker consists of a number of steps with the most crucial one in Association. To this end, we further introduce the concept of \textbf{pseudo-tracklet}\footnote{A tracklet is essentially an aggregation of a small number of consecutive sensor reports processed by a sensor level tracker~\cite{Oliver2002_Tracklet}. We use the tracklet as a short trajectory from a set of observations.}, constructed from the above pseudo-direction estimation. 
A pseudo-tracklet consists of a pair of vectors: $\{\{\widehat{\B{z}}_t\}_{t=1}^{T}, \{\widehat{\B{v}}_t\}_{t=2}^{T}\}$.
$\widehat{\B{z}}_{t}$ is an estimated observation with pseudo-direction $\widehat{\B{d}}_{T|T-\tau}$ and $\B{z}_T$ (Top of Fig.~\ref{fig:concept_association}), and $\widehat{\B{v}}_{t}$ is the forward direction linking between the estimated observations (Bottom of Fig.~\ref{fig:concept_association}).

The pseudo-tracklet can only be calculated from observations that are independent of the state of KF, and explicitly contains information about the movement of the object from the past to the present. We use this pseudo-tracklet to design the similarity metric in the association:
\begin{align}
    \label{eq:MCTrack}
    C^{\text{MCTrack}} &= \lambda C^{\text{angle}} + (1-\lambda) C^{\text{tracklet}},\\
    C^{\text{tracklet}} &= \frac{1}{T-1}\sum_{\tau=1}^{T-1}{\operatorname{GIoU}\LS{B_{\B{z}_{T-\tau}}, B_{\widehat{\B{z}}_{T-\tau}}}},\\
    C^{\text{angle}} &= \operatorname{GIoU}\LS{B_{\B{z}_{T}}, B_{\widehat{\B{x}}_{T\mid T-1}^{\B{R}}}},
\end{align}
where $\lambda$ is the weighting coefficient, $B$ represents the BBox with subscripts, and $\operatorname{GIoU}$~\cite{Rezatofighi2019_GIoU} denotes the similarity determined based on the distance between two BBoxes. 
In other words, $C^{\text{tracklet}}$ and $C^{\text{angle}}$ represent the similarity between the similarity between the pseudo-tracklet and the trajectory of the KF, and the current observation $\B{z}_T$ and the rotated state $\widehat{\B{x}}_{T\mid T-1}^{\B{R}}$ of the KF, respectively.

As shown in top of Fig.~\ref{fig:concept_association}, $C^{\text{tracklet}}$ directly correlates the observations $\B{z}_t$s of the KF trajectory with the estimated observations $\widehat{\B{z}}_t$ with the pseudo-direction. This approach, unlike the conventional method of correlating with only one observation value in the current frame, is more robust to motion.
The effectiveness of using both $C^{\text{tracklet}}$ and $C^{\text{angle}}$ is reported in Section~\ref{sec:ablation_study}.
Refer to Algorithm~\ref{algo:tracker} in Appendix~\ref{sec:training_inference_SIRA} for the pseudo-code of SIRA in inference.

In addition, as shown in bottom of Fig.~\ref{fig:concept_association} which represents the calculation of $C^{\text{angle}}$, the predicted state $\widehat{\B{x}}_{T\mid T-1}$ with KF from the previous state $\widehat{\B{x}}_{T-1}$ is rotated with a rotation matrix $\B{R}$ of angle $\phi_{\text{ave}}$. It can be calculated as $\B{p}_{\widehat{\B{x}}_{T\mid T-1}^{\B{R}}} = \B{R}(\B{p}_{\widehat{\B{x}}_{T\mid T-1}} - \B{p}_{\widehat{\B{x}}_{T-1}}) + \B{p}_{\widehat{\B{x}}_{T-1}}$,
where the angle $\widehat{\phi}_{\mathrm{ave}}$ can be calculated as $\widehat{\phi}_{\mathrm{ave}} = \frac{1}{T-2} \sum_{\rho=0}^{T-3} \widehat{\phi}_{T-\rho}$ such that $\widehat{\phi}_{T-\rho} = \cos ^{-1}\frac{\LS{\widehat{\B{v}}_{T-\rho} \cdot \widehat{\B{v}}_{T-\rho-1}}}{\left\|\widehat{\B{v}}_{T-\rho}\right\|\left\|\widehat{\B{v}}_{T-\rho-1}\right\|}$.
By using this rotated state $\widehat{\B{x}}_{T\mid T-1}^{\B{R}}$, we can avoid a high correlation between the predicted state assuming linear motion and the incorrect observation $\B{z}_T^{\text{noise}}$.

Our approach exploits the proposition that the temporally enhanced features across multiple frames from ETR allows for more robust estimation of the pseudo-direction $\widehat{\B{d}}_{T \mid T-\tau}$ from past frame $T-\tau$ to current frame $T$, compared with conventional single-frame based approaches. 

\subsection{Learning}
\label{sec:loss}
A loss function is constructed not only to acquire conventional detection capabilities, but also to provide a clear guideline to enhance tracking performance. 
It consists of two components: a loss between the predicted and the ground truth BBox ($\C{L}_{t}^{\text{BBox}}$), and a loss of the pseudo-direction in which an object has moved between frames and the actual movement direction ($\C{L}^{\text{DEst}}_{t}$), as shown in Fig.~\ref{fig:concept_architecture}.
\begin{equation} \label{eq:loss}
    \C{L}_{\theta} := \sum_{t=1}^{T}\LS{\C{L}^{\text{DEst}}_{t} + \C{L}_{t}^{\text{BBox}}}.
\end{equation}
For each training step, our training procedure calculates $\C{L}_{\theta}$ and does the backward for both $t=1$ to $t=T$ and $t=T$ to $t=1$ simultaneously.
Therefore, optimization $\min_{\theta}{\C{L}_{\theta}}$ can be viewed as a bidirectional backward-forward training through $T$ frames.
For more clear trainig procedure, refer to Fig.~\ref{fig:concept_SIRA_train_inference} in Appendix~\ref{sec:training_inference_SIRA}.

\paragraph{Oriented Bounding Box Loss:}
We pick the object’s center coordinates from the heatmap, and learn its attributes from feature representations through regression.
Regression functions, which are heatmap loss $\C{L}^{\text{h}}_{t}$, width \& Length loss $\C{L}^{\text{b}}_{t}$, orientation loss $\C{L}^{\text{r}}_{t}$, and offset loss $\C{L}^{\text{o}}_{t}$, compose the training objective by a linear combination:
\begin{equation}
    \label{eq:bbox_loss}
    \C{L}_{t}^{\text{BBox}} = \frac{1}{N_{\text{gt}}}\sum_{k=1}^{N_{\text{gt}}}\LS{\C{L}^{\text{b}}_{t,k}\!+\!\C{L}^{\text{r}}_{t,k}\!+\!\C{L}^{\text{o}}_{t,k}} - \frac{1}{N}\sum_{i=1}^{N}{\C{L}^{\text{h}}_{t,i}},    
\end{equation}
where $N$ denotes the total number of pixels in the heatmap and $N_{\text{gt}}$ is the total number of ground truth bounding boxes. Refer to~Appendix~\ref{sec:details_of_BBox_loss} for mathematical definition of each loss component. 

\paragraph{Pseudo-Direction Estimation Loss:}
$\C{L}^{\text{DEst}}$ represents a pseudo-direction estimation loss:
\begin{align}
\label{eq:DEst_lost}
    \C{L}^{\text{DEst}}_{t} &= \frac{1}{N_{\text{gt}}}\sum_{k=1}^{N_{\text{gt}}}{\C{L}^{\text{DEst}}_{t, k}},\\
\label{eq:DEst_lost_2}
    \C{L}^{\text{DEst}}_{t, k} &= \frac{1}{T-1} \sum_{\tau=1}^{T}\begin{cases}S_{L_1}\!\!\LS{\normx{\widehat{\B{d}}_{t \mid \tau}-\B{d}_{t \mid \tau}^{\text{gt}}}} & \tau \neq t \\ 0 & \tau = t\end{cases},
\end{align}
where $\widehat{\B{d}}_{t \mid \tau} = \C{G}_{\theta}^{\text{DEst}}\LS{\B{Z}_{t},\B{Z}_{\tau}}\LL{\B{p}_{t,k}^{\text{gt}}}$ denotes a two-dimensional direction from a position of time $\tau$ to a position of time $t$ as mentioned in Section~\ref{sec:tracking}. $\B{p}_{t,k}^{\text{gt}}$ denotes the coordinate $\LS{x_{t,k}, y_{t,k}}$ of the center of $k$-th ground truth object and $S_{L_1}\!\!\LS{\cdot}$ is a smooth $L_1$ loss~\cite{Girshick2015_RCNN}.
$\B{d}_{t \mid \tau}^{\text{gt}} = \B{p}_{t,k}^{\text{gt}} - \B{p}_{\tau,k}^{\text{gt}}$ denotes the ground truth direction, which can be calculated from the difference between the coordinates of the $k$-th object.
This loss improves the consistency of the detection positions between frames, which impacts both the detection and the tracking performance. 

\section{Experiments}
\label{sec:experiments}

\subsection{Experimental Setup}
\label{sec:experimental_setup}
Due to page limitations, more details on experimental settings are shown in  Appendix~\ref{sec:details_experimental_settings}.  

\paragraph{Dataset:}

We use the automotive radar dataset: \textit{Radiate}~\cite{Sheeny2021_RADIATE} in our experiments, the same as TempoRadar in~\cite{Li2022_TemporalRelations}.
The reasons to use this dataset are that it contains high-resolution radar images, provides well-annotated oriented bounding boxes with tracking IDs for objects, and records various real driving scenarios in adverse weather, please refer to Appendix~\ref{sec:dataset} for more details of the reasons.
\textit{Radiate} consists of video sequences recorded in adverse weathers, including sun, night, rain, fog and snow.
We follow the official 3 splits: ``train in good weather'' (22383 frames, only in good weather, sunny or overcast), ``train good \& bad weather'' (9749 frames, both good \& bad weather conditions), and ``test'' (11305 frames, all kinds of weather conditions).

\paragraph{Implementation:}
Our baseline detectors include: 1) RetinaNet~\cite{Lin2017_RetinaNet}, 2) CenterPoint~\cite{Xingyi2019_CenterPoint}, 3) BBAVectors~\cite{Yi2021_BBAVectors}, 4) TempoRadar~\cite{Li2022_TemporalRelations} (referred to as TR in all results).
We also implemented 
5) a Sequential TempoRadar (SeTR) that stacks self-attention for two consecutive frames and sequentially connects them through $T$ frames. We defer the description of the SeTR to Appendix~\ref{sec:SeTR}. We use ResNet-18 and ResNet-34 for the backbone feature extraction.

For MOT, we implemented several trackers that have been well demonstrated in this task for comparison.
These trackers include the following: CenterTrack~\cite{Zhou2020_CenterTrack} and OC-SORT~\cite{Cao2023_OCSORT}.
For the results of CenterTrack with TempoRadar and ResNet, we copied directly from the paper~\cite{Li2022_TemporalRelations} except for TempoRadar with 34 layers.
And for the KF-based method, we use the specific parameters and show the parameters in Appendix~\ref{sec:fundamentals_MOT}.
We follow~\cite{Sheeny2021_RADIATE} and exclude pedestrians and groups of pedestrians from detection and tracking targets, since only very few reflections are observed in these two kinds of objects.
For all numerical results, we apply a center crop with size $256\times 256$ upon input images and exclude the targets outside this scope. 
We additionaly report the detection results with the full size ($1152\times 1152$) images in Appendix~\ref{sec:additional_ablation_study}.

\paragraph{Metrics:}
We adopt the mean average precision (mAP) with intersection over union (IoU) at $0.3$, $0.5$, and $0.7$ (reported in Appendix~\ref{sec:additional_ablation_study}) to evaluate detection performance. The numbers
are averaged over 10 random seeds.
For MOT, we adopt MOTA~\cite{milan2016_mot16} and IDF1~\cite{Luiten2021_HOTA} as the main metrics. MOTA focuses more on the detection performance, while IDF1 reflects on the performance of association and identity preservation. Other metrics~\cite{milan2016_mot16} such as ID switch (IDs), fragmentation (frag), MT, and PT are also reported. Definitions of these MOT metrics are included in Appendix~\ref{sec:MOTMetric}. 

\subsection{Main Results}
\label{sec:result}

\paragraph{Detection:}

\begin{table}[t]
    \caption{Experimental results of object detection on \textit{Radiate}. The number following the model name indicates the \# of layers in the Resnet, and the number in parentheses indicates the \# of frames $T$.}
    \footnotesize
    \vspace{-2mm}
    \label{tab:results_detection}
    \setlength\tabcolsep{1.6pt}
    {
    \begin{tabular}{lccccc}
    \toprule

     & \multicolumn{2}{c}{\scriptsize{\textbf{Train good weather}}} & & \multicolumn{2}{c}{\scriptsize{\textbf{Train good \& bad weather}}}\\
    \cmidrule{2-3}\cmidrule{5-6}
     & mAP@0.3 & mAP@0.5 && mAP@0.3 & mAP@0.5 \\
    
    \midrule
    
    RetinaNet-18 (1) & 52.50\tiny{$\pm$1.81} & 37.83\tiny{$\pm$1.82} && 49.44\tiny{$\pm$1.32} & 31.57\tiny{$\pm$1.54} \\
    CenterPoint-18 (1) & 58.69\tiny{$\pm$3.09} & 49.41\tiny{$\pm$2.94} && 55.83\tiny{$\pm$3.28} & 44.48\tiny{$\pm$3.19} \\
    BBAVectors-18 (1) & 59.38\tiny{$\pm$3.47} & 50.53\tiny{$\pm$2.07} && 56.84\tiny{$\pm$3.45} & 45.43\tiny{$\pm$2.87} \\
    TR-18 (2) & 62.79\tiny{$\pm$2.01} & 53.11\tiny{$\pm$1.96} && 58.87\tiny{$\pm$3.31} & 46.42\tiny{$\pm$3.24} \\
    \midrule
    TR-18 (4) & 66.37\tiny{$\pm$1.62} & 53.23\tiny{$\pm$1.67} && 65.10\tiny{$\pm$1.67} & 52.47\tiny{$\pm$1.21} \\   
    SeTR-18 (4) & 65.97\tiny{$\pm$2.03} & 55.79\tiny{$\pm$2.12} && 64.62\tiny{$\pm$1.79} & 51.78\tiny{$\pm$1.81} \\
    \textbf{SIRA-18 (4)} & \cellcolor{gray!20}\textbf{67.28}\tiny{$\pm$1.47} & \cellcolor{gray!20}\textbf{56.98}\tiny{$\pm$1.35} && \cellcolor{gray!20}\textbf{65.37}\tiny{$\pm$1.76} & \cellcolor{gray!20}\textbf{52.88}\tiny{$\pm$1.60} \\    

    \midrule
    RetinaNet-34 (1) & 50.79\tiny{$\pm$3.10} & 35.61\tiny{$\pm$3.35} && 48.09\tiny{$\pm$3.85} & 31.10\tiny{$\pm$3.37} \\
    CenterPoint-34 (1) & 59.42\tiny{$\pm$1.92} & 50.17\tiny{$\pm$1.91} && 53.92\tiny{$\pm$3.44} & 42.81\tiny{$\pm$3.04} \\
    BBAVectors-34 (1) & 60.88\tiny{$\pm$1.79} & 51.26\tiny{$\pm$1.99} && 55.87\tiny{$\pm$2.90} & 44.61\tiny{$\pm$2.57} \\
    TR-34 (2) & 63.63\tiny{$\pm$2.08} & 54.00\tiny{$\pm$2.16} && 56.18\tiny{$\pm$4.27} & 43.98\tiny{$\pm$3.75} \\
    \midrule
    TR-34 (4) & 67.48\tiny{$\pm$0.94} & 57.01\tiny{$\pm$1.03} && 64.60\tiny{$\pm$2.08} & 51.99\tiny{$\pm$1.94} \\    
    SeTR-34 (4) & 67.30\tiny{$\pm$1.80} & 56.61\tiny{$\pm$1.83} && 65.51\tiny{$\pm$1.52} & 52.43\tiny{$\pm$1.51} \\
    \textbf{SIRA-34 (4)} & \cellcolor{gray!20}\textbf{68.68}\tiny{$\pm$1.12} & \cellcolor{gray!20}\textbf{58.11}\tiny{$\pm$1.40} && \cellcolor{gray!20}\textbf{66.14}\tiny{$\pm$0.83} & \cellcolor{gray!20}\textbf{53.79}\tiny{$\pm$1.14} \\
    
    \bottomrule
    \end{tabular}
    \vspace{-4mm}
    }
\end{table}

We report the detection results in Table \ref{tab:results_detection}. The benefits of exploiting longer temporal relation for radar object detection are evident in improvements of about $+3$ mAP$@0.3$ and about $+2.5$ mAP$@0.5$ from single frame of RetinaNet, CenterPoint, BBAVectors to two frames of the TempoRadar, and further more of about $+5$ mAP$@0.3$ and about $+4$ mAP$@0.5$ from two frames to four frames of the best among TempoRadar, SeTR, and SIRA. In both training splits, our SIRA consistently outperforms TempoRadar and its simple extension SeTR with $4$ radar frames. The improvement margin is more significant in the ``good \& bad weather" training split when ResNet34 is the backbone network.
We report the effectiveness of increasing the number of frames in Appendix~\ref{sec:additional_ablation_study}.

\paragraph{Tracking:} 
\begin{table}[t]
  \caption{Experimental results of MOT on \textit{Radiate}. 
  }
  \footnotesize
  \vspace{-2mm}
  \begin{threeparttable}[h]
  \centering
  \setlength\tabcolsep{1pt}
  {
  \begin{tabular}{lcccccc}
    \toprule
    
    \textbf{Train good weather} & MOTA$\uparrow$ &  IDF1$\uparrow$ & IDs$\downarrow$ & Frag.$\downarrow$ & MT$\uparrow$ & PT$\uparrow$ \\
    
    \midrule
    ResNet-18 (1) CenterTrack   & 13.01 &  -     & 873  & 920 & 269 & 254 \\
    ResNet-34 (1) CenterTrack   & 14.55 &  -     & 802  & 831 & 282 & 279 \\
    TR-18 (2) CenterTrack       & 33.59 &  -     & 349  & 498 & 145 & 330 \\
    TR-34 (2) CenterTrack       & 37.85 &  39.90 & 457  & 511 & 108 & 246 \\
    TR-34 (2) OC-SORT           & 40.74 & 45.01 & \cellcolor{gray!20}\textbf{151} & \cellcolor{gray!20}\textbf{291} & 124 & 172 \\
    \midrule
    TR-18 (4) CenterTrack       & 42.77 &  44.91 & 519  & 520 & 244 & \cellcolor{gray!20}\textbf{336} \\
    TR-34 (4) CenterTrack       & 43.64 &  44.17 & 503  & 538 & 197 & 326 \\
    TR-34 (4) OC-SORT           & 44.01 & 44.27 & 354 & 497 & 194 & 333 \\
    SeTR-18 (4) CenterTrack     & 42.11 &  50.33 & 658  & 561 & 261 & 317 \\
    SeTR-34 (4) CenterTrack     & 44.57 &  48.72 & 875  & 602 & 348 & 299 \\
    SeTR-34 (4) OC-SORT         & 40.16 & 28.20 & 775 & 689 & \cellcolor{gray!20}\textbf{370} & 305 \\
    \midrule
    ETR-34 (4) CenterTrack      & 46.06 &  50.81 & 1832 & 613 & 345 & 305 \\
    ETR-34 (4) OC-SORT          & 47.11 &  50.04 & 540  & 481 & 343 & 313 \\
    \textbf{SIRA-34 (4) CenterTrack}\tnote{*} & 47.30 & 50.16 & 1249 & 566 & 354 & 300 \\
    \textbf{SIRA-34 (4) OC-SORT} & \cellcolor{gray!20}\textbf{47.79} & \cellcolor{gray!20}\textbf{51.13} & 523 & 488 & 342 & 314 \\

    \bottomrule
    \end{tabular}
    \begin{tablenotes}
        \item[*] $C^{\text{tracklet}}$ is only used for association since this is not based on SORT.
    \end{tablenotes}
    \vspace{-2mm}
    }
\end{threeparttable}
\label{tab:track_main}
\vspace{-3mm}
\end{table}

Table~\ref{tab:track_main} illustrates the results of MOT. Similar conclusions can be made by observing the improvement margins in almost all metrics by using more radar frames. If we narrow down to the case of $4$ frames and with CenterTrack as the tracker, SIRA-34 shows a significant improvement of $+3.66$ over TR-34 and $+2.83$ over SeTR-34 in MOTA. The combination of SIRA+OC-SORT can further improve the MOT by another $+0.49$ over SIRA+CenterTrack. 

Compared with ETR (without $\C{L}^{\text{DEst}}_{t}$ for training), SIRA shows consistent improvement in both MOTA and IDF1, highlighting the effectiveness of modeling consistency in object movement. For other metrics such as Frag., MT, and PT, SIRA shows fluctuating but close-to-the-best performance. 
Full results, including the effectiveness of increasing the number of frames and other indicators, are reported in Appendix~\ref{sec:additional_ablation_study} due to paper space limitations.

\paragraph{Visualization:}

\begin{figure*}[t]
    \centering
    \includegraphics[width=\textwidth]{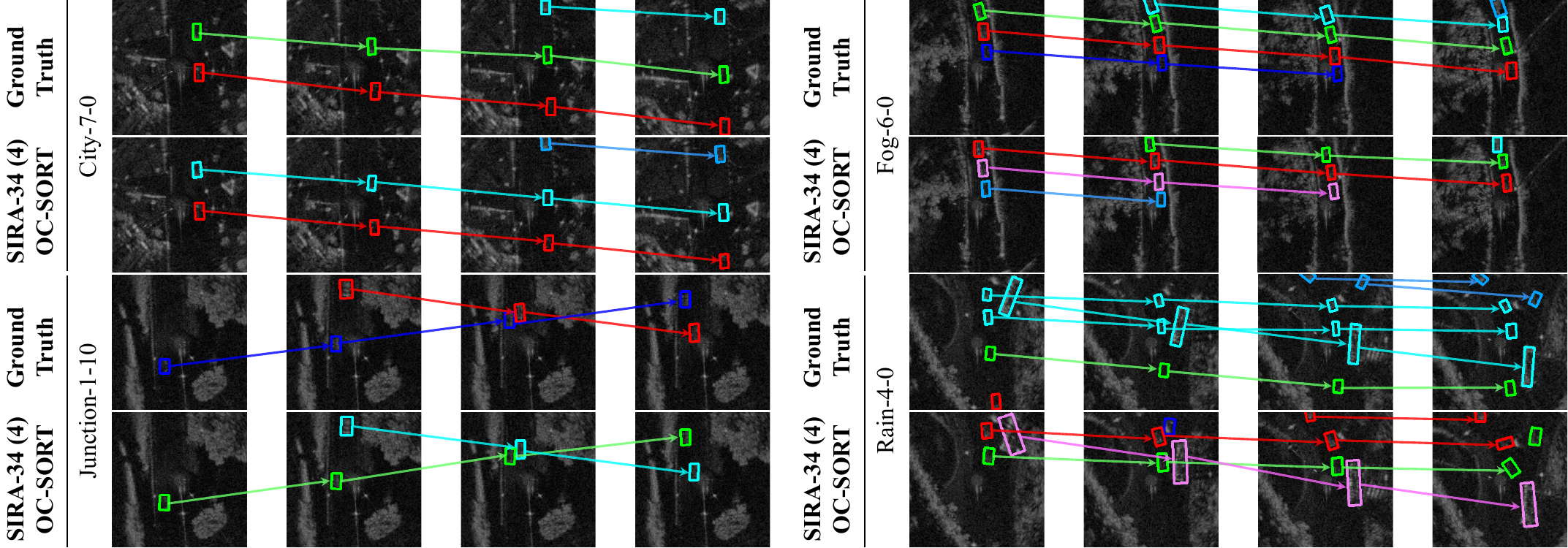}
    \vspace{-6mm}
    \caption{
        Visualizations on radar perception on \textit{Radiate}. $4$ sets of MOT results are shown in radar sequences of city-7-0, fog-6-0, junction-1-10 and rain-4-0. Each set contains $4$ frames. Bounding boxes are ground truth or object detection from SIRA. Colors indicate object IDs and plotted arrows show the motion of detected objects. 
    }
    \label{fig:results_vis_tracking_main}
\end{figure*}

We show the visualization results in Fig.~\ref{fig:results_vis_tracking_main}.
Each set of figures represents ground truth in the upper row and predictions in the lower row. It is observed that many of the predictions are made at approximately the same position as the annotations. Furthermore, correct predictions are observed for complex motions, including nonlinear motions.
More visualizations are included in Appendix~\ref{sec:additional_visualization_results} with more comparison to other baseline methods.

\subsection{Ablation Study}
\label{sec:ablation_study}

\begin{table*}[t]
    \footnotesize
    \centering
    \begin{subtable}[b]{0.21\linewidth}
        \centering
        \setlength\tabcolsep{2pt}
        \begin{tabular}{ccc}
        \toprule
        \textbf{$\mathcal{M}$} & mAP@0.3 & mAP@0.5 \\
        \midrule
        Mean & 65.15\scriptsize{$\pm$2.20} & 55.06\scriptsize{$\pm$2.07}  \\
        Sum & 65.76\scriptsize{$\pm$2.15} & 55.55\scriptsize{$\pm$1.59}  \\
        \textbf{Max} & \cellcolor{gray!20}\textbf{67.67}\scriptsize{$\pm$1.18} & \cellcolor{gray!20} \textbf{56.47}\scriptsize{$\pm$1.54}  \\
        \\
        \bottomrule
        \end{tabular}
        \caption{\textbf{Operations $\mathcal{M}$}. Using the Max operation works the best.}
        \label{tab:results_detection_abration_operation}
    \end{subtable}
    \hspace{1em}
    \begin{subtable}[b]{0.225\linewidth}
        \centering
        \setlength\tabcolsep{2pt}
        \begin{tabular}{cccc}
        \toprule
        $H_1$ & $H_2$ & mAP@0.3 & mAP@0.5   \\
        \midrule
        1 & 1 & 66.95\scriptsize{$\pm$1.47} & 56.65\scriptsize{$\pm$2.38} \\
        2 & 1 & 67.59\scriptsize{$\pm$0.83} & 57.59\scriptsize{$\pm$0.84} \\
        1 & 2 & 68.36\scriptsize{$\pm$0.94} & \cellcolor{gray!20}\textbf{58.46}\scriptsize{$\pm$0.91} \\
        2 & 2 & \cellcolor{gray!20} \textbf{68.68}\scriptsize{$\pm$1.12} & 58.11\scriptsize{$\pm$1.40} \\
        \bottomrule
        \end{tabular}
        \caption{\# \textbf{of MCAs}. A larger $H_2$ contributes more than a larger $H_1$.}
        \label{tab:results_detection_abration_number_of_MCA}
    \end{subtable}
    \hspace{1em}
    \begin{subtable}[b]{0.205\linewidth}
        \centering
        \setlength\tabcolsep{4pt}
        \begin{tabular}{ccc}
        \toprule
        $L$ & mAP@0.3 & mAP@0.5 \\
        \midrule
        1 & 68.68\scriptsize{$\pm$1.12} & 58.11\scriptsize{$\pm$1.40} \\
        2 & 68.68\scriptsize{$\pm$0.83} & 58.24\scriptsize{$\pm$1.19} \\
        3 & 69.12\scriptsize{$\pm$1.32} & \cellcolor{gray!20}\textbf{58.28}\scriptsize{$\pm$1.34} \\
        4 & \cellcolor{gray!20}\textbf{69.16}\scriptsize{$\pm$1.06} & 58.26\scriptsize{$\pm$1.27} \\
        \bottomrule
        \end{tabular}
        \caption{\# \textbf{of Stages}. More stages slightly improves the detection.}
        \label{tab:results_detection_abration_number_of_stages}
    \end{subtable}
    \hspace{1em}
    \begin{subtable}[b]{0.215\linewidth}
        \centering
        \setlength\tabcolsep{2pt}
        \begin{tabular}{cccc}
        \toprule
        $C^{\text{tracklet}}$ & $C^{\text{angle}}$ & MOTA$\uparrow$ & IDF1$\uparrow$ \\
        \midrule
        - & - & 47.11 & 50.04 \\
        \multicolumn{1}{c}{\checkmark} & - & 47.11 & 50.02 \\
        - & \multicolumn{1}{c}{\checkmark} & 47.00 & 50.05 \\
        \multicolumn{1}{c}{\checkmark} & \multicolumn{1}{c}{\checkmark} & \cellcolor{gray!20}\textbf{47.79} & \cellcolor{gray!20}\textbf{51.13} \\
        \bottomrule
        \end{tabular}
        \caption{\textbf{Associations $C$}. Using both $C^{\text{tracklet}}$ and $C^{\text{angle}}$ works the best.}
        \label{tab:track_association_ablation}
    \end{subtable}

    \caption{\textbf{SIRA ablation experiments} on \textit{Radiate}. If not specified, we used SIRA-34 (4) trained on train good weather and followed the experimental settings for other parameters.
    The best performance is marked in gray.}
    \label{tab:sira_ablation}
\end{table*}

\paragraph{Patch Merging Operator:}
In the context of patch merging within ETR, it is essential to merge feature vectors from overlapping positions. 
Multiple merging operations, including Mean, Sum and Max, can be considered. In the experiment, we use ETR-34 (4) as the model.
Table~\ref{tab:results_detection_abration_operation} shows the detection performance.
It is seen that the Max operation works best as the Mean and Sum operations may change the temporally enhanced features. We use the Max operation as the default.

\paragraph{Number of Masked MCA ($H_1$ and $H_2$):}
We investigated the effect of the number of masked MCA $H_1$ in TWA and $H_2$ of TRWA. 
The result in Table~\ref{tab:results_detection_abration_number_of_MCA} shows that larger $H$ improves the detection performance. More masked MCAs $H_2=2$ in the TRWA contributes to bigger improvement margin than using more masked MCAs $H_1=2$. We set  $H_1=2$ and $H_2=2$ as the default.

\paragraph{Number of Stages ($L$):}
We investigated the effect of the number of stages $L$ of ETR. Table~\ref{tab:results_detection_abration_number_of_stages} evaluates the detection performance when $L$ varies from only $1$ to $4$. Stacking more ETR stages slightly improves the detection performance. 

\paragraph{Association in MCTrack:}
In Table~\ref{tab:track_association_ablation}, the ablation study on association reveals that using both $C^{\text{tracklet}}$ and $C^{\text{angle}}$ leads to improved tracking performance. 
These facts indicate that SIRA enforces the spatio-temporal consistency and can be effective to deal with nonlinear object motion across consecutive frames.
See Appendix~\ref{sec:additional_ablation_study} for detailed evaluation results on the performance of Pseudo-Direction estimation and on the differences in $\lambda$.

\section{Conclusion}
\label{sec:conclusion}
We overcame the limitations of radar for effective object detection and tracking in automotive perception by introducing the SIRA framework, which includes ETR and MCTrack. SIRA exploits joint spatio-temporal consistency across multiple frames and enables reliable predictions despite low frame rates and nonlinear motion. Our approach outperforms previous state-of-the-art by a big margin in both detection and tracking. 

\bibliography{aaai24}
\clearpage

\section{Related Work for Visual Tracking}
\label{sec:visual_tracking} 
In recent years, KF-based approaches have gained popularity in the context of visual tracking, and various extensions have been proposed~\cite{Zhang2022_ByteTrack,Du2023_StrongSORT,Cao2023_OCSORT, Wojke2017_DeepSORT,Pang2021_QDSL,Zhang2021_FairMOT,Cao2022_DST,Maggiolino2023_DeepOCSORT,Tetsutaro2022_MHT}, exemplified by SORT~\cite{Bewley2016_SORT}. SORT can achieve high tracking performance, but it relies on the assumption that objects have consistent linear motion in a short time, which requires continuous observations. Therefore, it can face challenges when objects exhibit occlusion or nonlinear motion, requiring a high frame rate. To overcome occlusion problems, ByteTrack~\cite{Zhang2022_ByteTrack} uses the similarity between tracklets and low-scoring detection boxes to recover the true objects and filter out background detections. OC-SORT~\cite{Cao2023_OCSORT} introduces object motion computed from pre- and post-occlusion time pairs to address occlusion and non-linear motion. 
Our proposed framework extends recent KF-based methods and learning-based approaches by assuming high-density radar detection points. It explicitly considers strong object-level consistency by using multiple frames to capture the nonlinear motion of objects.

\section{Related Work for Radar Datasets}
\label{sec:dataset}
If we categorize open radar datasets into the ones with sparse detection points, dense points and low-level heatmap, RADIATE\cite{Sheeny2021_RADIATE} is the largest dense-point dataset with bounding box and tracking ID labels for both detection and tracking as shown in Table~\ref{tab:dataset}. 
We will use these datasets for further evaluation in the future.

\section{TempoRadar \cite{Li2022_TemporalRelations}}
\label{sec:temporadar}

Our ETR generalizes TempoRadar~\cite{Li2022_TemporalRelations} into a long time horizon and shares several key building blocks such as the top-$K$ feature selector $\C{S}_K$ and the design of the (temporal) masking matrix $\B{M}$ in the masked multi-head attention (MCA).

\paragraph{Top-$K$ Feature Selector $\C{S}_K$:}

To exploit the feature-level temporal relation, TempoRadar introduces a temporal relation layer (TRL).
Given the extracted features  $\B{Z}_{t} := \C{F}_{\theta}\LS{{I_{{t},{t-1}}}}$ and 
$\B{Z}_{t-1} := \C{F}_{\theta}\LS{{I_{{t-1},{t}}}}$ from the encoder, where ${I_{{t-1},{t}}}$ concatenates two consecutive radar frames $I_{t-1}$ and $I_t$ along the channel dimension in the order of $\LS{t-1, t}$, the feature selector $\C{S}_K$ of \eqref{eq:topK_selector} is defined as:
\begin{align}
    \B{H}_t & = \C{S}_K\LS{\B{Z}_t}:= \B{Z}_{t}\left[P_{t}^{\text{pre-hm}} \right] , \nonumber \\
    \B{H}_{t-1} & = \C{S}_K\LS{\B{Z}_{t-1}} := \B{Z}_{t-1}\left[ P_{t-1}^{\text{pre-hm}} \right], \notag 
\end{align}
where $\B{H}_{t/t-1} \in \R^{C\times K}$ and $P_{t}^{\text{pre-hm}}$ is defined as the set of $(x, y)$ coordinates corresponding to the $K$ selected features,
\begin{equation}
    \label{eq:coordinate_set}
    P_{t}^{\text{pre-hm}} := \left\{\LS{x, y} \Mid \LM{\B{C}_t}_{xy} \geq \LM{\B{C}_t}_K\right\}, 
\end{equation}
where $\B{C}_t=\C{G}_{\theta}^{\text{pre-hm}}\LS{\B{Z}_{t}}$ maps the channel dimension of the feature map via a learnable feedforward neural network (FNN) module $\C{G}^{\text{pre-hm}}_{\theta}:\R^{C\times\frac{H}{s}\times\frac{W}{s}}\to\R^{1\times\frac{H}{s}\times\frac{W}{s}}$ into a scalar feature map for feature ranking, $\LM{\B{C}_t}_K$ stands for the $K$-th largest value in $\B{C}_t$ over the spatial space $\frac{H}{s}\times \frac{W}{s}$, and the subscript $xy$ takes value at the coordinate $\LS{x,y}$.

\begin{table}[t]
    \caption{A list of open radar datasets in the format of dense points.}
    \footnotesize
    \centering
    \setlength\tabcolsep{1.5pt}
    {\begin{tabular}{lcccccc}
        \toprule
        \textbf{Dataset} & \textbf{\# of data} & \textbf{Radar format} & \textbf{BBox} & \textbf{Tracking ID} \\
        \midrule
        RADIATE~\cite{Sheeny2021_RADIATE} & 44K & dense points & 2D & \checkmark \\
        Zendar~\cite{Mostajabi2020_ZENDAR} & 4.8K & dense points & 2D & \checkmark \\
        TJ4DRadSet~\cite{Zheng2022_TJ4DRadSet} & 7.7K & dense points & 2D & \checkmark \\
        RADIal~\cite{Rebut2022_RADIal} & 25K & heatmap$+$points & 2D &  \\
        \bottomrule
    \end{tabular}
    }
    \label{tab:dataset}
\end{table}

\paragraph{Design of $\B{M}$ in Masked MCA:}
Let us stack the top-$K$ selected features from the two consecutive radar frames as 
$\B{H}_{t,t-1} := \LM{\B{H}_{t}, \B{H}_{t-1}}^{\top} \in \R^{2K\times C}$. The masked MCA takes $\B{H}_{t,t-1}$ and applies cross-frame attention over the two sets of features, as shown in Fig.~\ref{fig:TR_architecture}.

Since the position is lost in $\B{H}_{t,t-1}$, we generate the position information of the selected top-$K$ features via a learnable positional encoding network $\C{E}_{\theta}$ from the coordinate set $P_{t}^{\text{pre-hm}}$ of \eqref{eq:coordinate_set}
\begin{align}
\B{P}_{t}^{\text{enc}}=\C{E}_{\theta}\LS{P_{t}^{\text{pre-hm}}} \in\R^{K\times D_{\text{pos}}}, \notag 
\end{align}
where $D_{\text{pos}}$ is the dimension of positional encoding. We then supplement the positional encoding into feature vectors
\begin{align}
\B{H}_{t,t-1}^{\text{pos}}=\LM{\B{H}_{t,t-1}, \B{P}_{t,t-1}^{\text{enc}}}\in\R^{2K\times \LS{C+D_{\text{pos}}}}, \notag 
\end{align}
where $\B{P}_{t,t-1}^{\text{enc}}=\LM{\B{P}_{t}^{\text{enc}}, \B{P}_{t-1}^{\text{enc}}}^{\top} \in \R^{2K\times D_{\text{pos}}}$, and pass it to the masked MCA for temporal attention. 

In computing the temporal relation, the masked MCA follows \cite{Raffel2020_T2tUnifiedTransformer,Bao2020_UNILMv2,Hu2018_RelationNetworks,Hu2019_LocalRelationNetworks} and uses a temporal inductive bias with a masking matrix $\B{M}$:
\begin{equation}
    \C{A}\LS{\B{V}, \B{X}} := \;\operatorname{softmax}\LS{\frac{\B{M} +  q\LS{\B{X}} k\LS{\B{X}}^{\top}}{\sqrt{d}}} v\LS{\B{V}},\nonumber
\end{equation}
where $q\LS{\cdot}$, $k\LS{\cdot}$ and $v\LS{\cdot}$ are linear transformation layers and are referred to as query, keys and values, respectively, and $d$ is the dimension of the query and the keys.  For the temporal attention over $\{t, t-1\}$, we have $\C{A}\{\B{H}_{t,t-1}, \B{H}_{t,t-1}^{\text{pos}}\}$ with $\B{V} =\B{H}_{t,t-1}$ for the value and $\B{X}=\B{H}_{t,t-1}^{\text{pos}}$ for the key and query. The masking matrix $\B{M}$ is given as
\begin{equation}
    \B{M} := \LL{
    \begin{array}{l}
        \BB{I}_{K}, \B{1}_{K} \\
        \B{1}_{K}, \BB{I}_{K}
    \end{array}}
    + \sigma\LS{\LL{
    \begin{array}{l}
        \B{1}_{K}, \B{0}_{K} \\
        \B{0}_{K}, \B{1}_{K}
    \end{array}
    } - \BB{I}_{2K}},
\end{equation}
where $\BB{I}_{K}$ is the identity matrix of size $K$, $\B{1}_{K}$ and $\B{0}_{K}$ are the all-one and all-zero matrix with size $K \times K$, respectively, and $\sigma$ is a large negative constant, e.g., $-10^{10}$, to guarantee a near-zero value in the output through the softmax function. 
It can be shown that diagonal blocks in $\B{M}$ disable attention between features within the same frame, while off-diagonal blocks allow for cross-frame attention. The masked MCA may repeat multiple times. 

\section{Details of BBox Loss}
\label{sec:details_of_BBox_loss}

We pick the object’s center coordinates from the heatmap, and learn its attributes from feature representations through regression.
Regression functions, which are heatmap loss $\C{L}^{\text{h}}_{t}$, width \& Length loss $\C{L}^{\text{b}}_{t}$, orientation loss $\C{L}^{\text{r}}_{t}$, and offset loss $\C{L}^{\text{o}}_{t}$, compose the training objective by a linear combination as (\ref{eq:bbox_loss}):
\begin{equation}
    \C{L}_{t}^{\text{BBox}} = \frac{1}{N_{\text{gt}}}\sum_{k=1}^{N_{\text{gt}}}\LS{\C{L}^{\text{b}}_{t,k}\!+\!\C{L}^{\text{r}}_{t,k}\!+\!\C{L}^{\text{o}}_{t,k}} - \frac{1}{N}\sum_{i=1}^{N}{\C{L}^{\text{h}}_{t,i}},\nonumber
\end{equation}
where $N$ denotes the total number of pixels in the heatmap and $N_{\text{gt}}$ is the total number of ground truth bounding boxes. 
Each loss is as follows:
\begin{align}
    \C{L}^{\text{h}}_{t,i} = & \mathds{1}_{\LM{c_{t,i}=1}}\LS{1-\widehat{c}_{t,i}}^{\alpha} \log \LS{\widehat{c}_{t,i}}\nonumber\\ 
    &+ \mathds{1}_{\LM{c_{t,i} \neq 1}}\LS{1-c_{t,i}}^{\beta}{\widehat{c}_{t,i}}^\alpha \log \LS{1-\widehat{c}_{t,i}},
\end{align}
where $c_{t,i}$ and $\widehat{c}_{t,i}$ denote the ground-truth and predicted value at $i$-th coordinate in $\C{G}_{\theta}^{\text{hm}}\LS{\B{Z}_{t}^{\mathrm{hm}}}$, and $\alpha$ and $\beta$ are hyper-parameters and are chosen empirically with 2 and 4, respectively.
\begin{align}
    \C{L}^{\text{b}}_{t,k} = & S_{L_1}\!\!\LS{\normx{\C{G}_\theta^{b}\!\LS{\B{Z}_{t}\!\!\LL{P_{t,k}^{\text{gt}}}}\!-\!\LS{w_{t,k}, h_{t,k}}^{\top}}},\\
    \C{L}^{\text{r}}_{t,k} = & S_{L_1}\!\!\LS{\normx{\C{G}_\theta^{r}\!\LS{\B{Z}_{t}\!\!\LL{P_{t,k}^{\text{gt}}}}\!\!-\!\LS{\cos{\vartheta_{t,k}}, \!\sin{\vartheta_{t,k}}}\!\!^{\top}}}\!,\\
    \C{L}^{\text{o}}_{t,k} = & S_{L_1}\!\!\LS{\normx{\C{G}_\theta^{o}\!\LS{\B{Z}_{t}\!\!\LL{P_{t,k}^{\text{gt}}}}\!-\!\LS{o_{x,t,k}, o_{y,t,k}}^{\top}}},
\end{align}
where $P_{t,k}^{\text{gt}}$ denotes the coordinate $\LS{x_{t,k}, y_{t,k}}$ of the center of $k$-th ground truth object, $\LS{w_{t,k}, h_{t,k}}$ is the width \& length, and $\LS{o_{x,t,k}, o_{y,t,k}}$ is the offset as follows:
\begin{equation}
    \LS{o_{x,t,k}, o_{y,t,k}} = \LS{\frac{x_{t,k}}{s}-\round{\frac{x_{t,k}}{s}}, \frac{y_{t,k}}{s}-\round{\frac{y_{t,k}}{s}}}.    
\end{equation}

\section{Sequential TempoRadar (SeTR)}
\label{sec:SeTR}
One might postulate: ``What are the implications of extending \textit{TempoRadar} to cover more consecutive radar frames?'' The answer might be two-fold. On the one hand, one should expect improved performance under the assumption that most radar features are present over more than just $2$ frames, considering a typical radar frame rate of $>4$ fps (\textit{Radiate} dataset has $4$ fps ~\cite{Sheeny2021_RADIATE}). On the other hand, directly applying temporal attention to a longer time horizon incurs a quadratic computation complexity (refer to (\ref{eq:computational_complexity_TR})) over the number of features from each frame $K$ and the number of frames $T$. 

\begin{figure}[t]
    \centering
    \begin{minipage}{.48\textwidth}
        \centering
        \includegraphics[width=\textwidth]{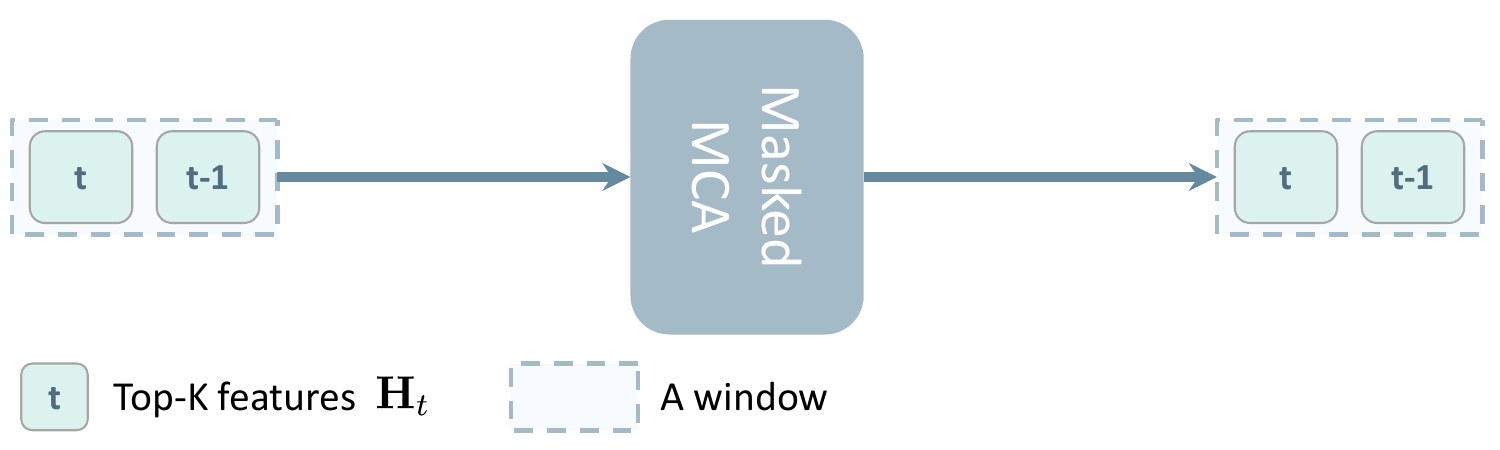}
        \subcaption{Masked MCA of TempoRadar.}
        \label{fig:TR_architecture}
    \end{minipage}
    \vspace{10pt}
    \\
    \begin{minipage}{.48\textwidth}
        \centering
        \includegraphics[width=\textwidth]{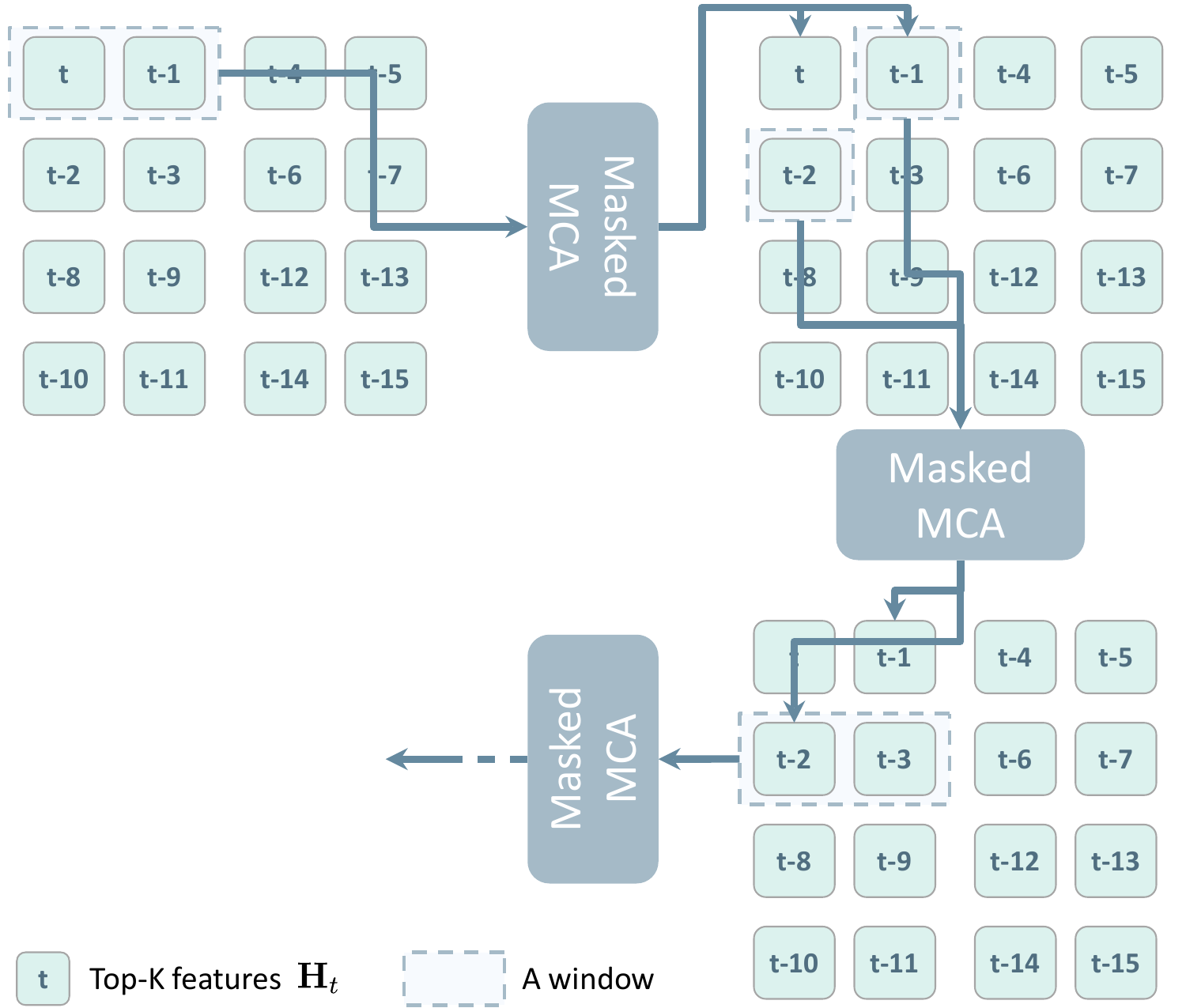}
        \subcaption{Masked MCA of SeTR.}
        \label{fig:SeTR_architecture}
    \end{minipage}
    \caption{Masked MCA. (a)~TempoRadar~\cite{Li2022_TemporalRelations} computes masked multi-head cross-attention (MCA) over the top-$K$ selected features from a time horizon of only $T=2$ consecutive radar frames. (b)~ SeTR computes masked MCA for two consecutive radar frames at a time, the same as the TempoRadar in (a), but slides the window of two frames after each MCA sequentially to cover a longer time horizon of $T > 2$ frames.}
\end{figure}

One straightforward way for a scalable TempoRadar is to stack temporal feature attention for two consecutive frames and sequentially connect them, which we refer to as \textit{sequential TempoRadar} (SeTR). As illustrated in Fig.~\ref{fig:SeTR_architecture}, SeTR computes masked MCA for two consecutive radar frames at a time, the same as the TempoRadar in Fig.~\ref{fig:TR_architecture}, but slides the window of two frames after each MCA sequentially to cover a longer time horizon of $T > 2$ frames.

\section{Training and Inference Pipelines for SIRA}
\label{sec:training_inference_SIRA}

\begin{figure*}[t]
    \centering
    \includegraphics[width=\textwidth]{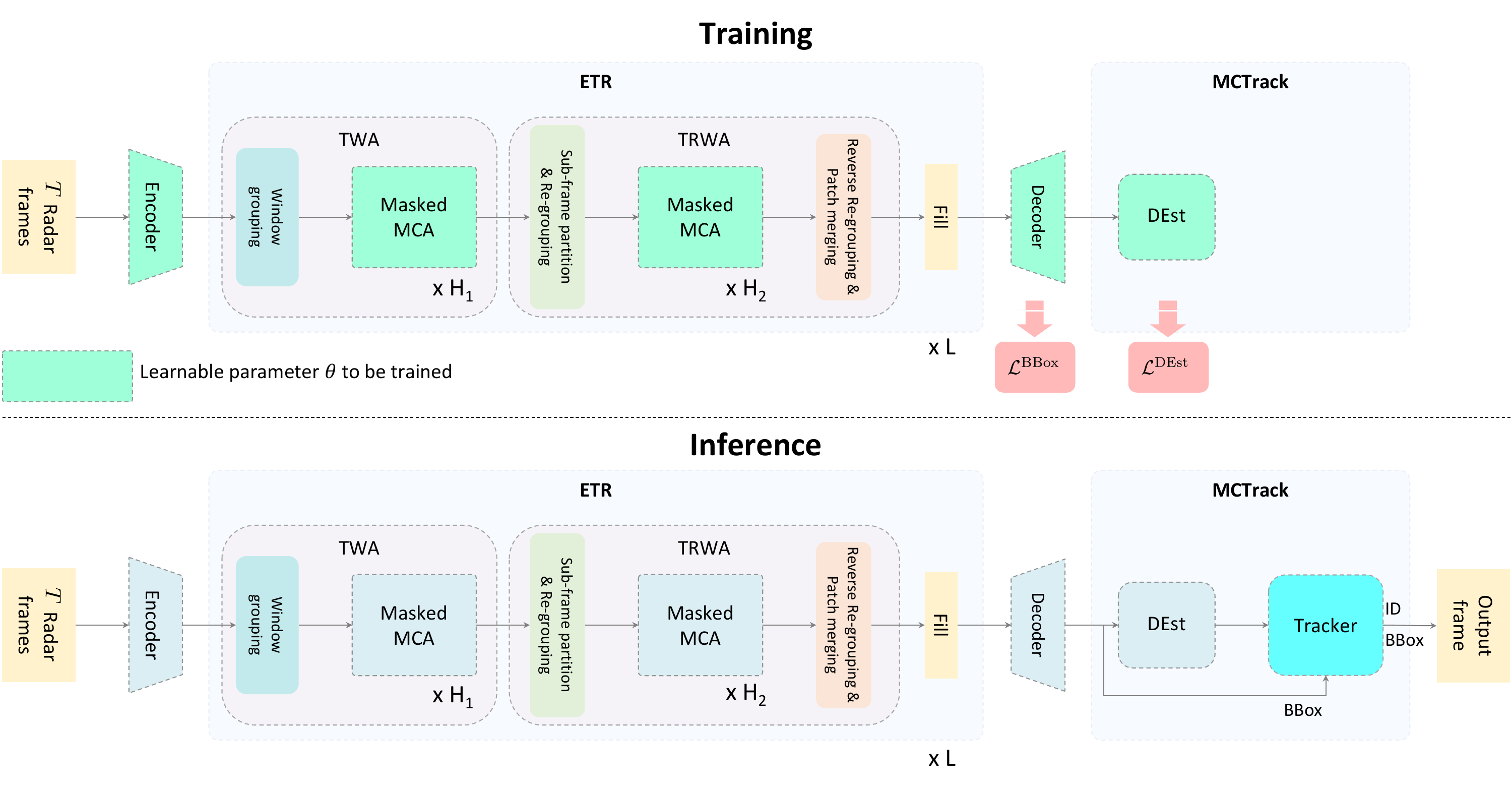}
    \vspace{-5mm}
    \caption{
        Training and Inference Pipelines for SIRA. (Top) SIRA takes $T$ consecutive radar frames into the training pipeline, and computes $\C{L}^{\text{BBox}}$ at the decoder and the pseudo-direction loss $\C{L}^{\text{DEst}}$ at the DEst module in \textbf{Motion Consistency for Training}  of Section~\ref{sec:tracking}. Learnable modules are hatched in \textcolor{lightgreen}{\textbf{light green}}.
        (Bottom) SIRA attaches a tracker at the DEst module output to further enforce \textbf{Motion Consistency for Inference} of Section~\ref{sec:tracking}. The tracker incorporates the motion similarity of \eqref{eq:MCTrack} for association. Learnable parameters are frozen during inference. 
    }
    \label{fig:concept_SIRA_train_inference}
\end{figure*}

\paragraph{Training Pipeline for SIRA:}
To train SIRA, we takes $T$ consecutive radar frames, pass them into the training pipeline in the Top diagram of Fig.~\ref{fig:concept_SIRA_train_inference}, and compute the  loss function $\C{L}^{\text{BBox}}$ of (\ref{eq:bbox_loss}) at the decoder output for detection loss
and the pseudo-direction loss $\C{L}^{\text{DEst}}$ of (\ref{eq:DEst_lost}) at the output of the DEst module (detailed in \textbf{Motion Consistency for Training} of Section~\ref{sec:tracking}). Through backpropagation, the learnable modules, hatched in \textcolor{lightgreen}{\textbf{light green}}, are updated using the derived loss value. 

\paragraph{Inference Pipeline for SIRA:}
In the bottom of Fig.~\ref{fig:concept_SIRA_train_inference}, we show the inference pipeline for SIRA. Noticeably, a tracker is attached to the DEst module to further enforce the motion consistency via the concept of pseudo-tracklet, detailed in \textbf{Motion Consistency for Inference} of Section~\ref{sec:tracking}.  All learnable parameters during training are frozen in the inference. We further include the pseudo-code of the inference pipeline in Algorithm~\ref{algo:tracker}.
A typical tracker consists of five steps: Prediction, Association, Update, Deletion, and Initialization. 
We can integrate our MCTrack with standard trackers (e.g., OC-SORT) by incorporating the key components (highlighted in \textcolor{codegreen}{\textbf{green}} in Algorithm~\ref{algo:tracker}), e.g., the use of motion similarity of \eqref{eq:MCTrack}, to the Association step.

\begin{algorithm}[!t]
    \SetAlgoLined
    \DontPrintSemicolon
    \SetNoFillComment
    \footnotesize
    \KwIn{A radar frame sequence $\texttt{V}$; encoder $\texttt{Enc}$; decoder $\texttt{Dec}$; object detector $\texttt{ETR}$; direction estimator $\texttt{DEst}$; detection score threshold {$\gamma$}; birth threshold {$\beta$}}
    
    \KwOut{Tracks $\C{T}$ of the video}
    
    Initialization: $\C{T} \leftarrow \emptyset$\;
    \For{frame $f_k$ in $\texttt{V}$}{
    	\tcc{Fig.\ref{fig:concept_architecture}, and Fig.\ref{fig:concept_SIRA_train_inference}}

        \BlankLine	
        \tcc{**\textbf{predict bboxes with ETR}**}
            \textcolor{codegreen}{$\C{F}_k \leftarrow \texttt{Enc}(f_k)$} \;
            \textcolor{codegreen}{$\C{F}_k \leftarrow \texttt{ETR}(\C{F}_k)$} \;
            \textcolor{codegreen}{$\C{D}_k \leftarrow \texttt{Dec}(\C{F}_k)$} \;

        \BlankLine	
        \tcc{**\textbf{tracking with MCTrack}**}
            \textcolor{codegreen}{$\C{J}_k \leftarrow \texttt{DEst}(\C{F}_k)$} \;
            $\C{D}_{high} \leftarrow \emptyset$ \;
            \textcolor{codegreen}{$\C{J}_{high} \leftarrow \emptyset$} \;
            \For{$d, j$ in $\C{D}_k, \C{J}_k$}{
        	\If{$d.score > \gamma$}{
                    $\C{D}_{high} \leftarrow  \C{D}_{high} \cup \{d\}$ \;
                    \textcolor{codegreen}{$\C{J}_{high} \leftarrow  \C{J}_{high} \cup \{j\}$} \;
        	}
    	}
    	
        \BlankLine	
        \BlankLine
    	\tcc{predict new locations of tracks}
    	\For{$t$ in $\C{T}$}{
                $t \leftarrow \texttt{KalmanFilter.predict}(t)$ \;
    	}
    	
        \BlankLine
        \BlankLine
    	\tcc{Fig.\ref{fig:concept_association}}
    	\tcc{association}
            \textcolor{codegreen}{
                Associate $\C{T}$ and $\C{D}_{high}\&\C{J}_{high}$ with \texttt{Similarity} Eq.\ref{eq:MCTrack} \;
            }
            $\C{D}_{remain} \leftarrow \text{remaining unmatched object from } \C{D}_{high}$ \;
            $\C{T}_{remain} \leftarrow \text{remaining matched tracks from } \C{T}$ \;

        \BlankLine	
        \BlankLine
    	\tcc{update status of matched tracks}
    	\For{$t$ in $\C{T}_{remain}$}{
                $t \leftarrow \texttt{KalmanFilter.update}(t)$ \;
    	}
    
        \BlankLine
        \BlankLine
    	\tcc{delete unmatched tracks}
    	$\C{T} \leftarrow \C{T} \setminus \C{T}_{remain}$ \;
    	
        \BlankLine
        \BlankLine
    	\tcc{initialize new tracks}
            \For{$d$ in $\C{D}_{remain}$}{
                \If{$d.score > \beta$}{
                    $\C{T} \leftarrow  \C{T} \cup \{d\}$ \;
        	}
    	}
    }
    Return: $\C{T}$
    \caption{Pseudo-code of SIRA for Inference.}
    \algorithmfootnote{In \textcolor{codegreen}{\textbf{green}} is the key of our method. }
    \label{algo:tracker}
    
\end{algorithm}

\begin{figure}[t]
    \centering
    \includegraphics[width=0.45\textwidth]{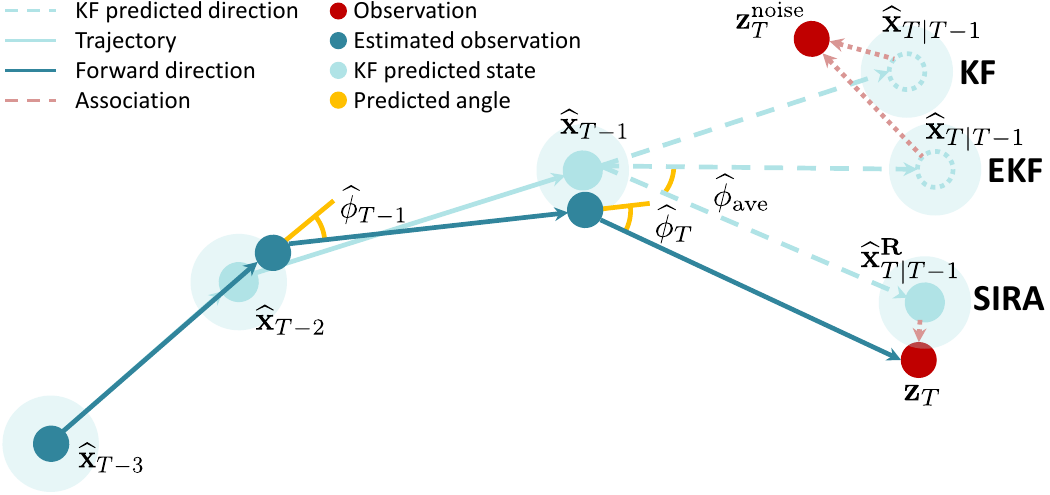}
    \caption{Trajectory prediction of SIRA with KF and EKF.}
    \label{fig:concept_tracking_KF}
    \vspace{-5mm}
\end{figure}

\paragraph{Extension with higher-order KF:}
As shown in Fig.~\ref{fig:concept_tracking_KF}, we expect SIRA can deal with nonlinear motion to an extent as the average predicted angle $\widehat{\phi}_{\text{ave}}$ can correct the KF predicted state $\widehat{\B{x}}_{T\mid T-1}$ closer to the right observation $\B{z}_T$. SIRA can be extended with higher-order KF (e.g., extended/unscented KF) to further improve the predicted state $\widehat{\B{x}}_{T\mid T-1}$ with a proper nonlinear model and the average predicted angle $\widehat{\phi}_{\text{ave}}$, yielding improved trajectory predictions. 

\paragraph{The choice of patch size:}
Since the patch is a subset of Top-$K$ features in each frame, we have the patch size $M \in [1, K]$. Given the number of frames $U$ in each window, the smaller the patch size $M$, the smaller the window size $UM$ in the re-grouping stage of the TRWA block (see the top right of Fig.~\ref{fig:ETR_architecture_block2} for an example of $U=4$ frames and $M=K/2$), and the lower the computational complexity of the window-based  attention which is quadratic with respect to $UM$. 
On the other hand, a small $M$ may limit the number of features to be correlated across windows and reduces the connectivity of temporal attention. 

\section{Details of Experimental Settings}
\label{sec:details_experimental_settings}

\paragraph{Dataset} 
\begin{figure*}[t]
    \centering
    \includegraphics[width=\textwidth]{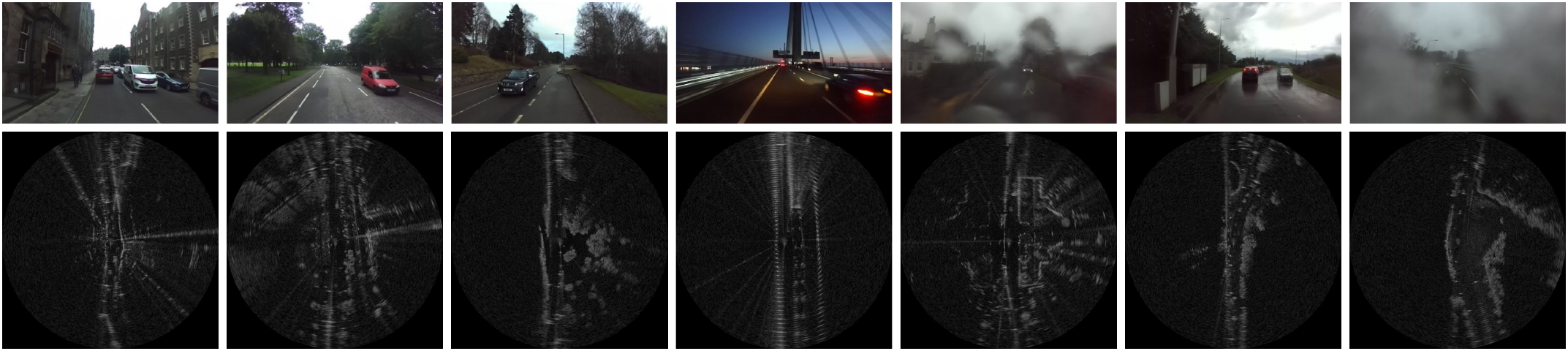}
    \vspace{-5mm}
    \caption{Visualization of RGB and corresponding radar frames. From left to right, the scenes are from City-3-7, City-7-0, Junction-1-10, Night-1-4, Fog-6-0, Rain-4-0 and Snow-1-0 in \textit{Radiate}. Albeit of more coarse-grained and less semantic features, radar frames are much more resilient than RGB frames in adverse weather and low lighting conditions. }
    \label{fig:vis_rgb_imgs}
    \vspace{-5mm}
\end{figure*}
To facilitate the research on robust and reliable vehicle perception, \textit{Radiate} dataset was collected in $7$ scenarios under various weather and lighting conditions: Sunny (Parked), Sunny/Overcast (Urban), Overcast (Motorway), Night (Motorway), Rain (Suburban), Fog (Suburban) and Snow (Suburban). It includes multiple sensor modalities from radar and optical images to 3D LiDAR point clouds and GPS.  $8$ object classes, i.e., car, van, truck, bus, motorbike, bicycle, pedestrian and group of pedestrian, were annotated on the radar frames.
The data format of radar frames generated from dense point clouds, where the pixel values indicate radar reflection magnitude. 
Radiate adopted the Navtech CTS350-X FMCW radar, a scanning radar that provides $360^{\circ}$ high-resolution range-azimuth BEV images at $4$ Hz. It was set to have 100-meter maximum operating range with a distance resolution of $0.175$ m, an azimuth resolution of $1.8^{\circ}$  and an elevation resolution of $1.8^{\circ}$. It does not provide Doppler information.
Radar frames in Cartesian are provided as .png at $1152 \times 1152$ resolution. Nearest neighbour interpolation was used to convert the radar framess from the polar coordinate to the Cartesian one. Each pixel in the Cartesian coordinate represents a grid of $0.17361 \times 0.17361 \text{m}^2$. In other words, the field of view is about $\LL{-100\text{m}, 100\text{m}}$ in one axis and $ \LL{-100\text{m}, 100\text{m}}$ in the other axis in BEV. Radiate dataset has official $3$ splits: ``train in good weather'' which consists of $31$ sequences ($22383$ frames, only in good weather, sunny or overcast), ``train good \& bad weather'' which consists of $12$ sequences ($9749$ frames, both good \& bad weather conditions), and ``test'' which consists of $18$ sequences ($11305$ frames, all kinds of weather conditions).
Fig.~\ref{fig:vis_rgb_imgs} shows sampled RGB and corresponding radar frames under adverse weather and low lighting conditions.
We separately train models on the former two training sets and evaluate on the test set. 

\paragraph{Hyper-parameters}
The hyper-parameters used in our experiments of Section~\ref{sec:experiments} are shown in Table~\ref{tab:exp_hyper_params}.
The table is divided into three parts, Data, Architecture, and Training, each with parameter names, notations, and values.
\begin{table}[t]
    \footnotesize
    \caption{Hyper-parameters used in our experiments.}
    \label{tab:exp_hyper_params}
    \setlength\tabcolsep{2.7pt}
    {
    \begin{tabular}{clcc}
    \toprule

    \multicolumn{2}{c}{\textbf{Name}} & \textbf{Notation} & \textbf{Value} \\
    \midrule
    \multirow{6}{*}{\rotatebox{90}{\textbf{Data}}}  & dataset & - & \textit{Radiate} \\
     & train good weather & - & 22383 \\
     & train good \& bad weather & - &  9749 \\
     & test & - & 11305 \\
     & cropped image size & $H \times W$ & 256 $\times$ 256 \\
     & full image size & $H \times W$ & 1152 $\times$ 1152 \\
    \midrule
     \multirow{11}{*}{\rotatebox{90}{\textbf{Architecture}}} & position dimention & $D_{\text{pos}}$ & 64 \\
     & downsampling ratio & $s$ & 4 \\
     & \# of top-$K$s & $K$ & 8 \\
     & \# of sets of top-$K$ & $U$ & 2 \\
     & \# of ETR stages & $L$ & 1 \\
     & \# of masked MCAs: TWA & $H_1$ & 2 \\
     & \# of masked MCAs: TRWA & $H_2$ & 2 \\
     & operation & $\C{M}$ & max \\
     & coefficient & $\lambda$ & 0.5 \\
     & detection score threshold & $\gamma$ & 0.08\\
     & birth threshold & $\beta$ & 0.20\\
    \midrule
    \multirow{9}{*}{\rotatebox{90}{\textbf{Training}}} & batch size & - & 16 \\
     & epoch & - & 10 \\
     & optimizer & - & Adam \\
     & learning rate & - & 5e-4 \\
     & schedule for train good $\times 0.1$ & - & 5 \\
     & schedule for train good \& bad $\times 0.1$ & - & 2 \\
     & weight decay for detection & - & 1e-2 \\
     & weight decay for tracking & - & 1e-5 \\
     & \# of GPUs & - & 1 \\
 
    \bottomrule
    \end{tabular}
    }
    \vspace{-5mm}
\end{table}

\begin{figure}[t]
    \centering
    \includegraphics[width=0.4\textwidth]{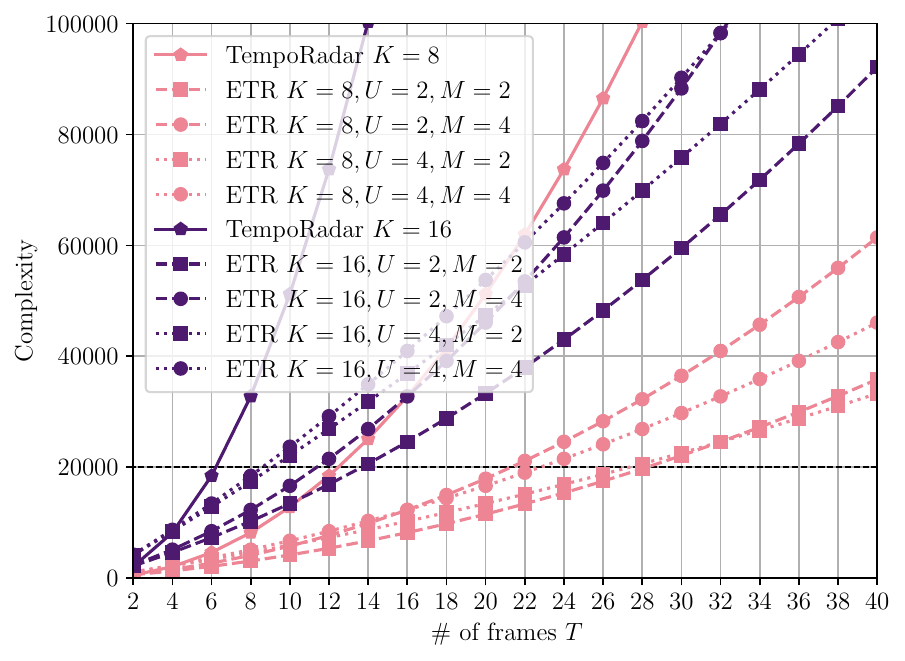}
    \caption{
        Comparison of computational complexities of TempoRadar (solid) and ETR (dashed) as a function of the number of frames $T$ (along the $x$-axis) and the number of selected features $K$ (grouped in different colors). 
    }
    \label{fig:computational_comlexity}
    \vspace{-5mm}
\end{figure}

\begin{table}[t]
    \footnotesize
    \caption{Additional results of object detection on \textit{Radiate} for mAP@0.7. The number following the model name indicates the \# of layers in the ResNet, and the number in parentheses indicates the \# of frames $T$.}
    \label{tab:results_detection_full_evaluation}
    \setlength\tabcolsep{5.4pt}
    {
    \begin{tabular}{lcc}
    \toprule
    
    \textbf{mAP@0.7} & \scriptsize{\textbf{Train good weather}} & \scriptsize{\textbf{Train good \& bad weather}} \\
    
    \midrule
    
    RetinaNet-18 (1) & 8.46\scriptsize{$\pm$0.61} &6.97\scriptsize{$\pm$1.24} \\
    CenterPoint-18 (1) & 19.02\scriptsize{$\pm$1.80}& 14.43\scriptsize{$\pm$2.56} \\
    BBAVectors-18 (1) & 19.72\scriptsize{$\pm$1.10}& 15.07\scriptsize{$\pm$1.76} \\
    TR-18 (2) & 20.57\scriptsize{$\pm$1.47}& 15.59\scriptsize{$\pm$2.31} \\
    
    TR-18 (4) & 19.59\scriptsize{$\pm$0.78}& 19.62\scriptsize{$\pm$1.33} \\   
    SeTR-18 (4) & 21.90\scriptsize{$\pm$1.12}& 19.65\scriptsize{$\pm$0.84} \\
    \textbf{SIRA-18 (4)} & \cellcolor{gray!20}\textbf{21.95}\scriptsize{$\pm$1.72}& \cellcolor{gray!20}\textbf{19.66}\scriptsize{$\pm$1.87} \\    

    \midrule
    RetinaNet-34 (1) & 7.67\scriptsize{$\pm$1.71}& 6.93\scriptsize{$\pm$1.60} \\
    CenterPoint-34 (1) & 18.93\scriptsize{$\pm$1.46}& 13.43\scriptsize{$\pm$1.92} \\
    BBAVectors-34 (1) & 19.86\scriptsize{$\pm$1.36}& 14.67\scriptsize{$\pm$1.45} \\
    TR-34 (2) & 21.08\scriptsize{$\pm$1.66}& 14.35\scriptsize{$\pm$2.15} \\

    TR-34 (4) & 22.46\scriptsize{$\pm$1.76}& 19.03\scriptsize{$\pm$1.10} \\    
    SeTR-34 (4) & 21.68\scriptsize{$\pm$1.24}& 19.63\scriptsize{$\pm$1.29} \\
    \textbf{SIRA-34 (4)} & \cellcolor{gray!20}\textbf{22.81}\scriptsize{$\pm$0.86}& \cellcolor{gray!20}\textbf{19.85}\scriptsize{$\pm$0.95} \\

    \bottomrule
    \end{tabular}
    }
\end{table}

\begin{table}[t]
    \footnotesize
    \caption{Comparison on object detection with full size images. Comparison on object detection with full size images.}
    \label{tab:results_detection_comparison_non_crop}
    \setlength\tabcolsep{11.5pt}
    {
    \begin{tabular}{lcc}
    \toprule
    
    \textbf{Train good weather} & mAP@0.3 & mAP@0.5 \\

    \midrule
    
    FasterRCNN-50 (1)~\cite{Sheeny2021_RADIATE} & - & 45.31 \\
    FasterRCNN-101 (1)~\cite{Sheeny2021_RADIATE} & - &  45.84 \\
    TR-18 (2)~\cite{Li2022_TemporalRelations} & - &  48.02 \\
    TR-34 (2)~\cite{Li2022_TemporalRelations} & - &  48.66 \\
    \midrule
    ETR-34 (4) & 65.10 & 49.19 \\
    \textbf{SIRA-34 (4)} & \cellcolor{gray!20}\textbf{65.67} & \cellcolor{gray!20}\textbf{51.49} \\
    \midrule
    ETR-34 (6) & 67.19 & 49.37 \\
    \textbf{SIRA-34 (6)} & \cellcolor{gray!20}\textbf{67.72} & \cellcolor{gray!20}\textbf{52.14} \\
    \midrule
    ETR-34 (8) & 65.53 & 50.59 \\    
    \textbf{SIRA-34 (8)} & \cellcolor{gray!20}\textbf{67.82} & \cellcolor{gray!20}\textbf{52.55} \\    
    \midrule
    ETR-34 (10) & 64.24 & 50.12 \\  
    \textbf{SIRA-34 (10)} & \cellcolor{gray!20}\textbf{66.03} & \cellcolor{gray!20}\textbf{50.77} \\
    \bottomrule
    \end{tabular}
    \vspace{-5mm}
    }
\end{table}
\begin{table*}
\begin{threeparttable}[h]
  \footnotesize
  \caption{Experimental results of multiple object tracking on \textit{Radiate}. The number following the model name indicates the \# of layers in the Resnet backbone, and the number in parentheses indicates the \# of frames $T$.}
  \label{tab:track_full_metrics}
  \centering
  \setlength\tabcolsep{8.9pt}
  {
  \begin{tabular}{lcccccccccc}
    \toprule
    
    \textbf{Train good weather} & MOTA$\uparrow$ & MOTP$\uparrow$ & IDF1$\uparrow$ & IDs$\downarrow$ & FP$\downarrow$ & FN$\downarrow$ & Frag.$\downarrow$ & MT$\uparrow$ & ML$\downarrow$ & PT$\uparrow$ \\
    
    \midrule
    ResNet-18 (1) CenterTrack & 13.01 & 70.26 & - & 873 & - & - & 920 & 269 & - & 254 \\
    ResNet-34 (1) CenterTrack & 14.55 & 70.05 & - & 802 & - & - & 831 & 282 & - & 279 \\

    TR-18 (2) CenterTrack & 33.59 & 73.49 & - & 349 & - & - & 498 & 145 & - & 330 \\
    TR-34 (2) CenterTrack & 37.85 & 71.85 & 39.90 & 457 & 970 & 6114 & 511 & 108 & 422 & 246 \\

    \midrule
    TR-18 (4) CenterTrack & 42.77 & 70.38 & 44.91 & 519 & 1061 & 5206 & 520 & 244 & 196 & 336 \\
    TR-34 (4) CenterTrack & 43.64 & 71.58 & 44.17 & 503 & 854 & 5892 & 538 & 197 & 253 & 326 \\
    SeTR-18 (4) CenterTrack & 42.11 & 68.71 & 50.33 & 658 & 1481 & 4672 & 561 & 261 & 198 & 317 \\
    SeTR-34 (4) CenterTrack & 44.57 & 71.65 & 48.72 & 875 & 1511 & 4606 & 602 & 348 & 129 & 299 \\

    \midrule
    ETR-34 (4) CenterTrack & 46.06 & 70.23 & 50.81 & 1832 & 1141 & 4904 & 613 & 345 & 126 & 305 \\
    ETR-34 (4) OC-SORT & 47.11 & 70.08 & 50.04 & 540 & 1411 & 4523 & 481 & 343 & 120 & 313 \\

    \textbf{SIRA-34 (4) CenterTrack}\tnote{*} & 47.30 & 70.19 & 50.16 & 1249 & 1218 & 4756 & 566 & 354 & 122 & 300 \\
    \textbf{SIRA-34 (4) OC-SORT} & 47.79 & 70.09 & 51.13 & 523 & 1408 & 4513 & 488 & 342 & 120 & 314 \\

    \bottomrule
    \end{tabular}
    \begin{tablenotes}
        \item[*] For CenterTrack, $C^{\text{tracklet}}$ is only used for association since this tracker is not based on SORT.
    \end{tablenotes}
    }
\end{threeparttable}
\end{table*}

\begin{table*}
  \footnotesize
  \vspace{-2mm}
  \caption{Ablation study of various number of frames for multiple object tracking on train good weather. We used SIRA-34 OC-SORT.}
  \vspace{-2mm}
  \label{tab:track_frames_full_metrics}
  \centering
  \setlength\tabcolsep{11.1pt}
  {
  \begin{tabular}{ccccccccccc}
    \toprule
    \# of frames $T$ & MOTA$\uparrow$ & MOTP$\uparrow$ & IDF1$\uparrow$ & IDs$\downarrow$ & FP$\downarrow$ & FN$\downarrow$ & Frag.$\downarrow$ & MT$\uparrow$ & ML$\downarrow$ & PT$\uparrow$ \\

    \midrule
    4 & 47.79 & 70.09 & 51.13 & 523 & 1408 & \cellcolor{gray!20}\textbf{4513} & 488 & \cellcolor{gray!20}\textbf{342} & \cellcolor{gray!20}\textbf{120} & 314 \\
    6 & \cellcolor{gray!20}\textbf{49.22} & \cellcolor{gray!20}\textbf{71.70} & \cellcolor{gray!20}\textbf{51.87} & \cellcolor{gray!20}\textbf{399} & \cellcolor{gray!20}\textbf{1032} & 4692 & \cellcolor{gray!20}\textbf{306} & 255 & 172 & \cellcolor{gray!20}\textbf{349} \\
    8 & 46.12 & 69.55 & 50.21 & 487 & 1076 & 4746 & 449 & 312 & 139 & 325 \\

    \bottomrule
  \end{tabular}
  \vspace{-3mm}
  }
\end{table*}

\section{Comparison of Complexity Analysis}
\label{sec:complexity_analysis}

Fig.~\ref{fig:computational_comlexity} compares the computational complexity of TempoRadar in (\ref{eq:computational_complexity_TR}) and ETR in (\ref{eq:computational_complexity_ETR}) as a function of the number of consecutive radar frames $T$, under two settings of the number of selected features $K=8$ and $K=16$ (grouped in two different colors). For each setting of $K$, we further include four ETR variants (denoted by different markers) with different combinations of hyper-parameters of the number of consecutive radar frames  within one temporal window $U$ and the number of features within one patch $M$. Under both settings, ETR provides more affordable temporal attention over longer time horizons $T$ than TempoRadar.  

While (\ref{eq:computational_complexity_ETR}) represents the complexity of the general ETR module, \cite{Yataka2024_SCTR} presents a special case of ETR with $U=2$, $M=K/2$ and a stride $K/4$ (a special sub-frame partition with $50\%$ overlapping). For this special case, the computational complexity is shown to be $2 K^2\LS{3T-4} L$. 

\section{Definition of MOT Metrics}
\label{sec:MOTMetric}

We adopt the series of MOT metrics~\cite{milan2016_mot16,Luiten2021_HOTA} for evaluation. We pick several key metrics in the experiments: MOTA (Multiple Object Tracking Accuracy), IDF1, ID switch (IDs), track fragmentations (Frag.), mostly tracked (MT), and partially tracked (PT). The MOTA score is calculated by
\begin{equation*}
    \text{MOTA} = 1 - \frac{\sum_t (\text{FN}_t + \text{FP}_t + \text{IDSW}_t)}{\sum_t \text{GT}_t},
\end{equation*}
where $t$ is the frame index, GT is the number of ground-truth objects, and FN and FP refer to false negative and false positive detection, respectively. The value of MOTA is in the range $(-\infty, 100]$. It can be deemed as the combination of detection and tracking performance, and is widely used as the main metric for accessing multiple object tracking quality. 

IDF1 evaluates the identity preservation ability and focuses on the association performance.
Specifically, IDF1 calculates a bijective (one-to-one) mapping between the sets of ground truth trajectories and predicted trajectories (unlike MOTA at the detection level) and is a function of 
\begin{itemize}
    \item{IDTPs} (identity true positives): the matches in the overlapping sections of trajectories that are correctly associated with the same identity;
    \item{IDFNs} (identity false negatives): instances where the ground truth has an identity that the prediction fails to identify. This often occurs in non-overlapping sections of matched trajectories or when the tracker loses track of an object; 
    \item{IDFPs} (identity false positives): instances where the prediction assigns an identity that does not exist in the ground truth. This often happens in the case of over-segmentation or incorrect identity assignments;
\end{itemize}

\begin{equation}
\text{IDF1}=\frac{2\text{IDTP}}{2\text{IDTP}+\text{IDFP}+\text{IDFN}}.
\end{equation}
The rest of these metrics all reflect the quality of predicted tracklets. For detailed definitions and calculations of MOT metrics, please refer to~\cite{milan2016_mot16}.

\begin{figure}[t]
    \centering
    \includegraphics[width=0.4\textwidth]{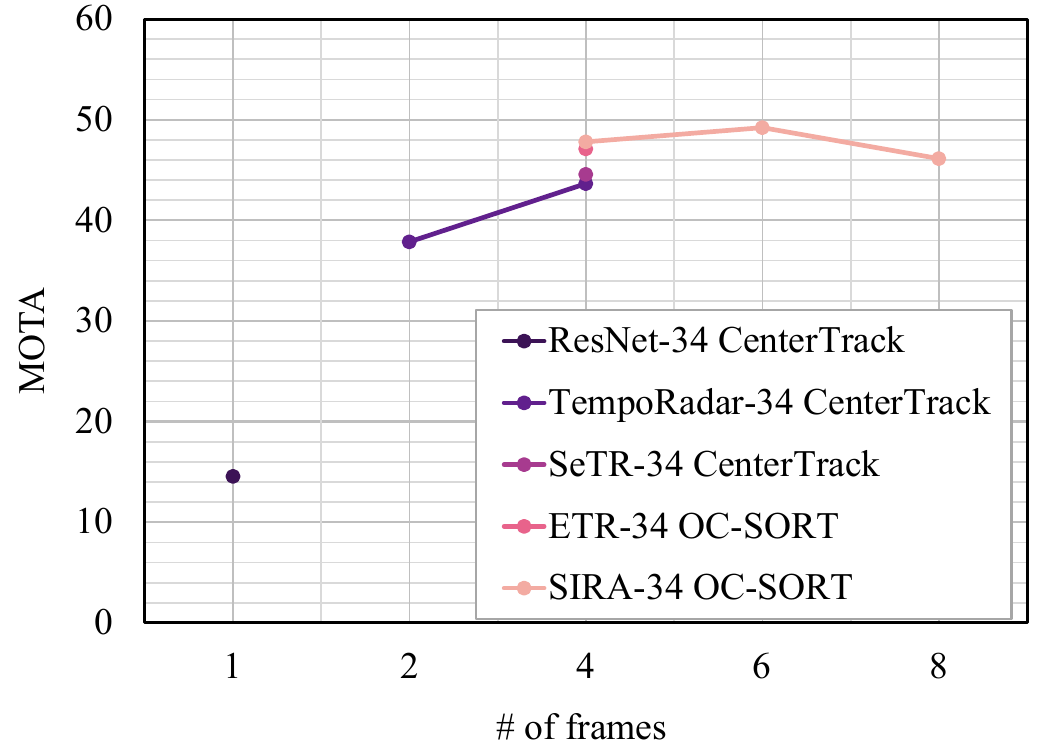}
    \caption{
        Tracking performance as a function of number of frames $T$. Compared with the single-frame baseline (ResNet-34 CenterTrack), SIRA with $T=6$ consecutive frames results in a margin of $+34.67$ MOTA. 
    }
    \label{fig:comparison_graph}
    \vspace{-2mm}
\end{figure}
\begin{figure}[t]
    \centering
    \includegraphics[width=0.4\textwidth]{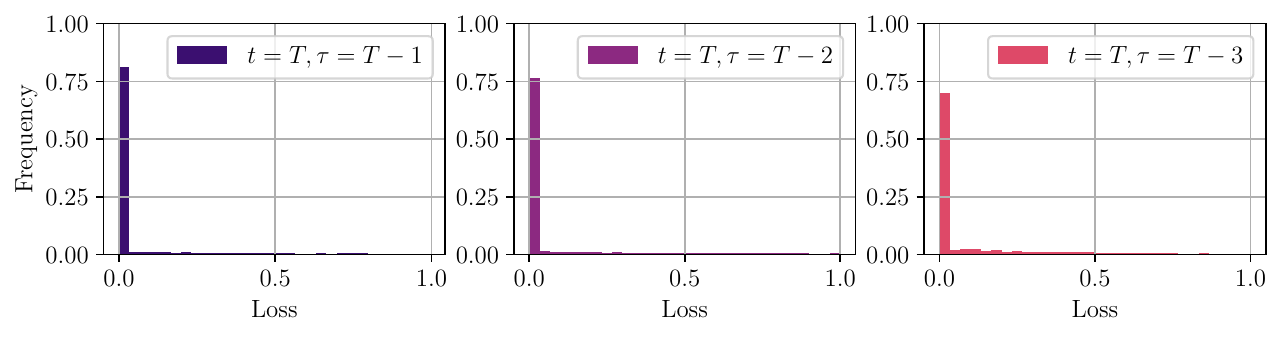}
    \caption{Pseudo-direction smooth $L_1$ loss in different time steps. }
    \vspace{-5mm}
    \label{fig:tracking_loss}
\end{figure}
\begin{figure}[t]
    \centering
    \includegraphics[width=0.3\textwidth]{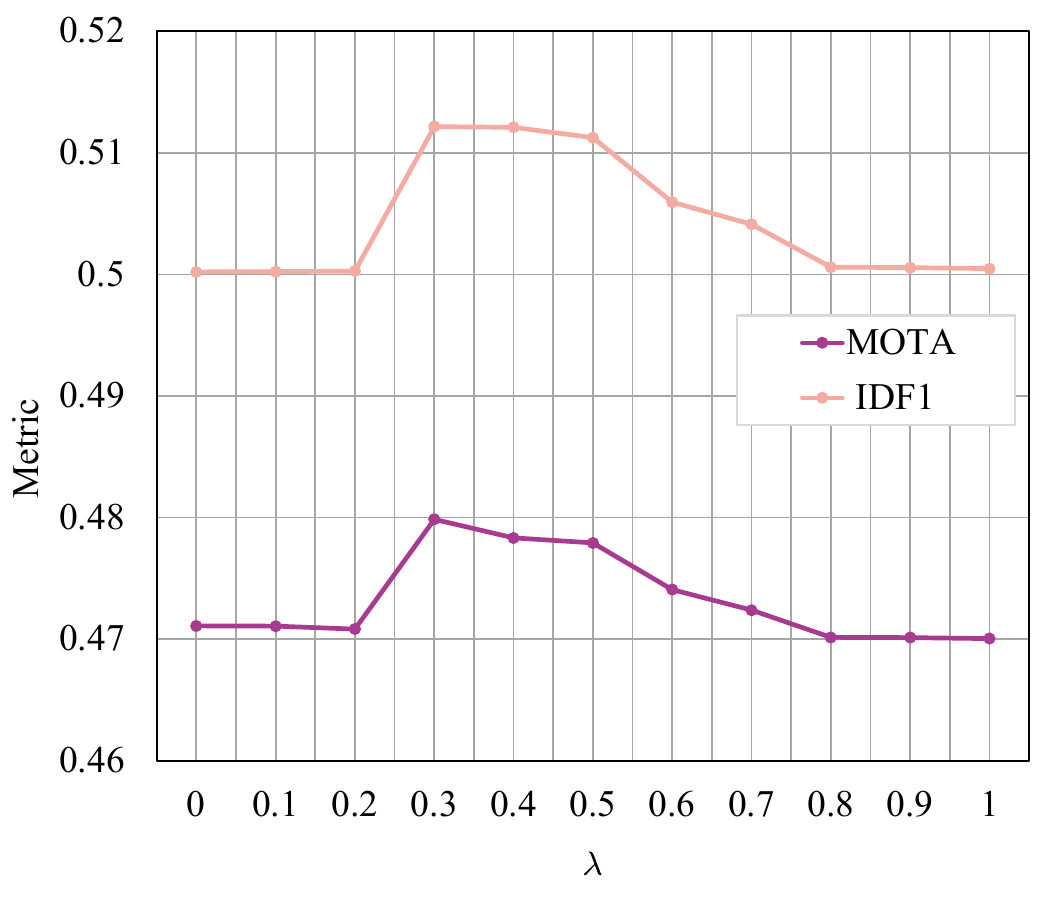}
    \caption{Performance variation due to different parameter $\lambda$.}
    \label{fig:comparison_lambda}
    \vspace{-7mm}
\end{figure}
\begin{figure*}[t]
    \centering
    \includegraphics[width=\textwidth]{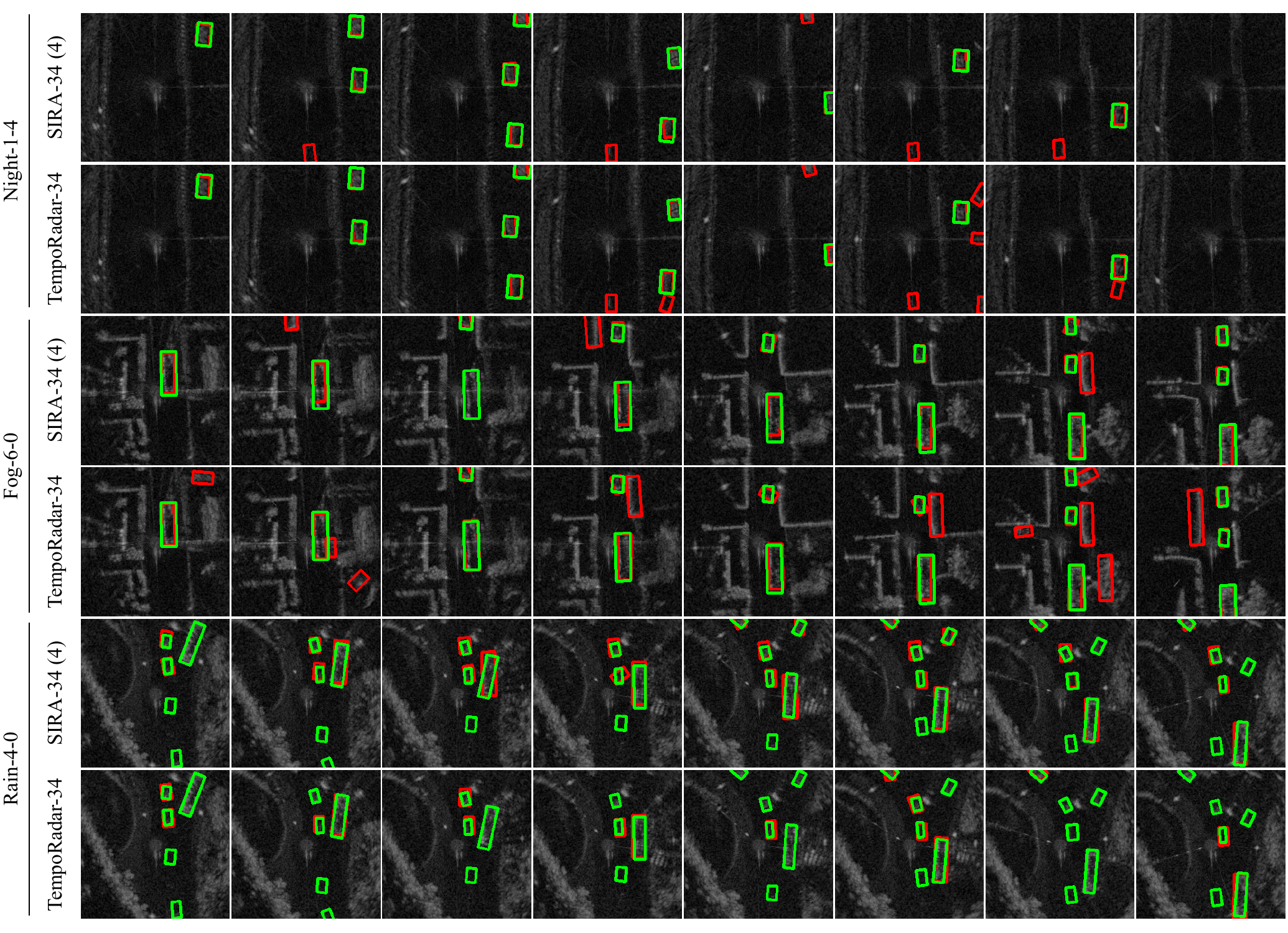}
    \caption{
        \textbf{Sampled detection results with cropped radar frames} on three scenarios: Night-1-4 (Top $2$ Rows), Fog-6-0 (Middle $2$ Rows) and Snow-4-0  (Bottom $2$ Rows) on \textit{Radiate}. For each scenario, we compare the SIRA and TempoRadar. Green boxes represent ground truth and red ones are predictions. The column represents consecutive radar frames. TempoRadar shows more false positives (FNs as unpaired red boxes) than SIRA, particularly in the first two scenarios. 
    }
    \label{fig:results_vis_detection_1}
\end{figure*}

\begin{figure*}[t]
    \centering
    \includegraphics[width=\textwidth]{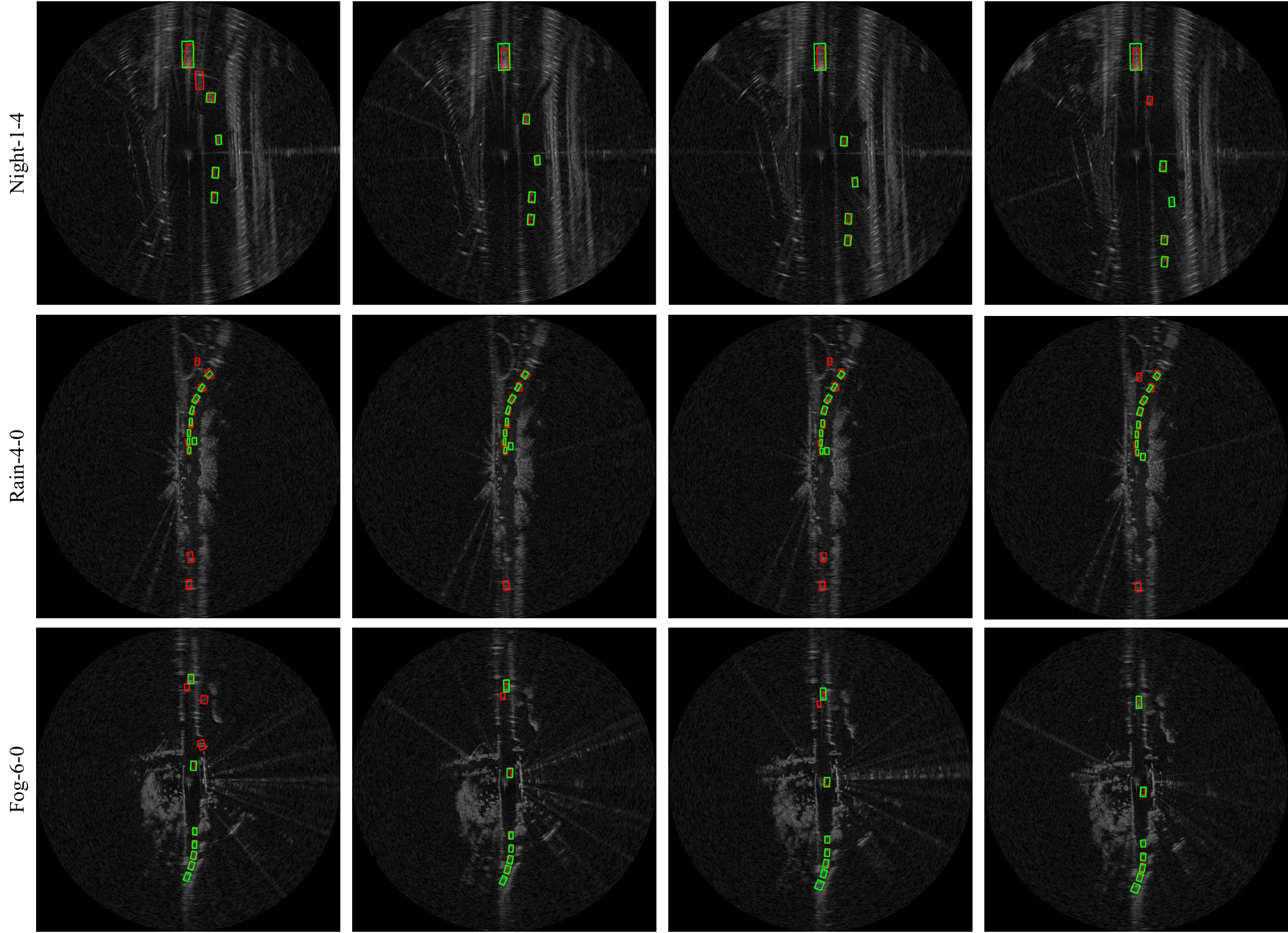}
    \caption{\textbf{Sampled detection results of SIRA-34 (6) with full size radar frames} on three scenarios: Night-1-4 (Top), Rain-4-0  (Middle) and Fog-6-0  (Bottom ) on \textit{Radiate}. Green boxes represent ground truth and red ones are predictions. The column represents consecutive radar frames.}
    \label{fig:vis_imgs_non_crops}
\end{figure*}
\begin{figure*}[t]
    \centering
    \includegraphics[width=\textwidth]{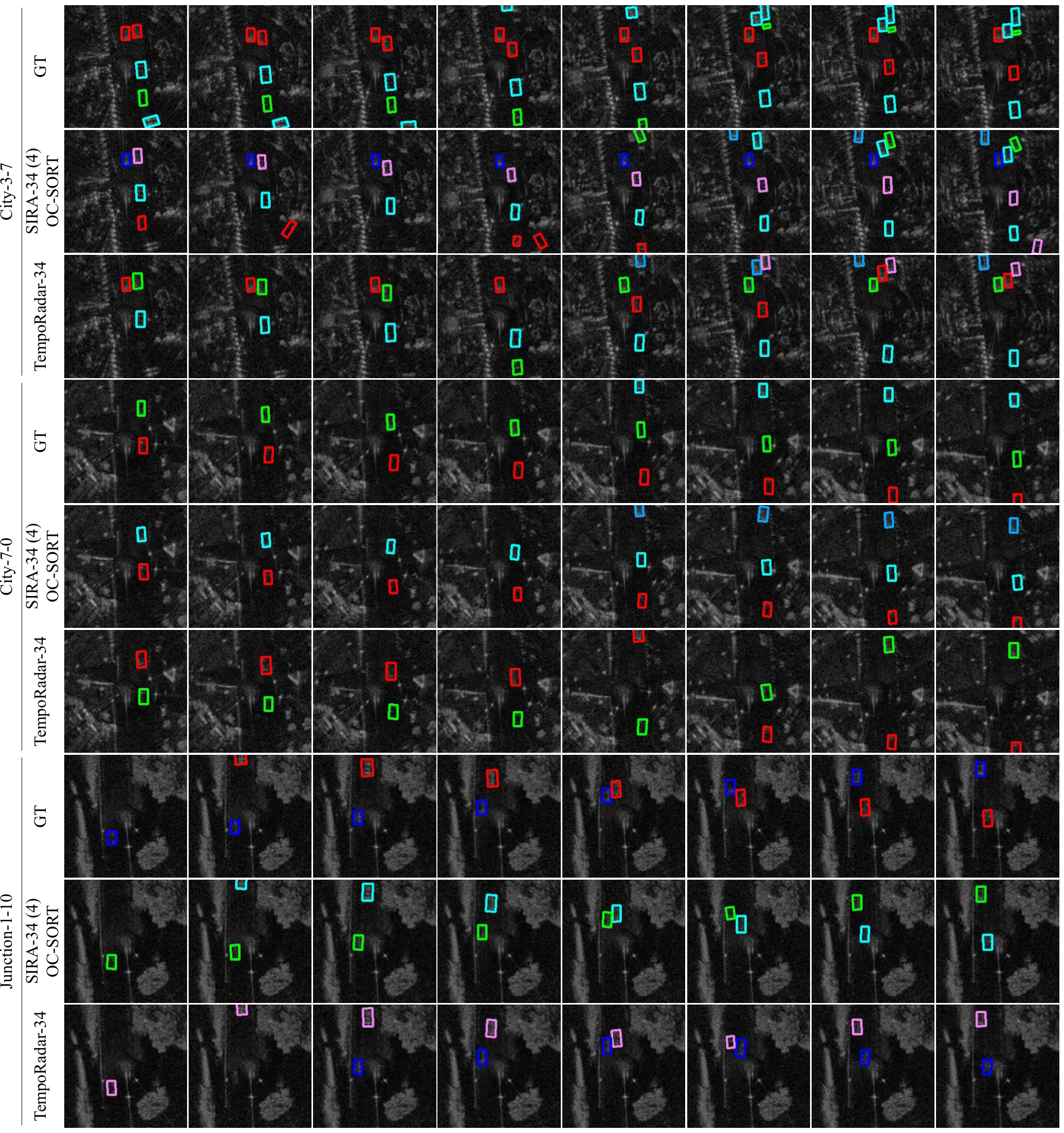}
    \caption{
          \textbf{Sampled tracking results} on three scenarios: City-3-7 (Top $3$ Rows), City-7-0  (Middle $3$ Rows) and Junction-1-10 (Bottom $3$ Rows) on \textit{Radiate}. For each scenario, we include ground truth (GT), SIRA (SIRA-34 (4)) and TempoRadar (TempoRadar-34). The color of bounding boxes represents the object ID. The column represents consecutive radar frames. 
    }
    \label{fig:results_vis_tracking_1}
\end{figure*}

\begin{figure*}[t]
    \centering
    \includegraphics[width=\textwidth]{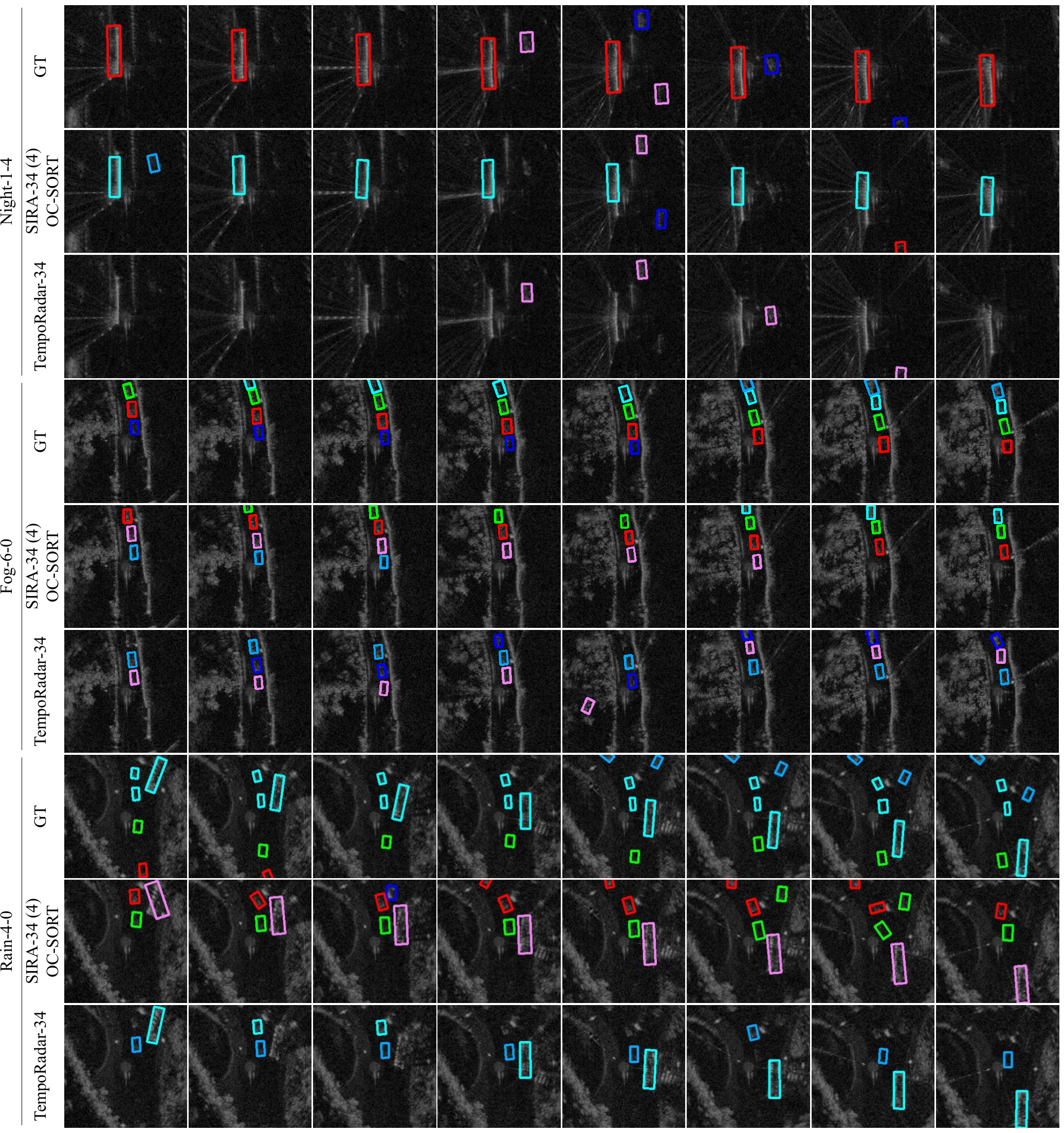}
    \caption{
        \textbf{Sampled tracking results} on three scenarios: Night-1-4 (Top $3$ Rows), Fog-6-0   (Middle $3$ Rows) and Rain-4-0 (Bottom $3$ Rows) on \textit{Radiate}. For each scenario, we include ground truth (GT), SIRA (SIRA-34 (4)) and TempoRadar (TempoRadar-34). The color of bounding boxes represents the object ID. The column represents consecutive radar frames. 
    }
    \label{fig:results_vis_tracking_2}
\end{figure*}

\begin{figure*}[t]
    \centering
    \includegraphics[width=\textwidth]{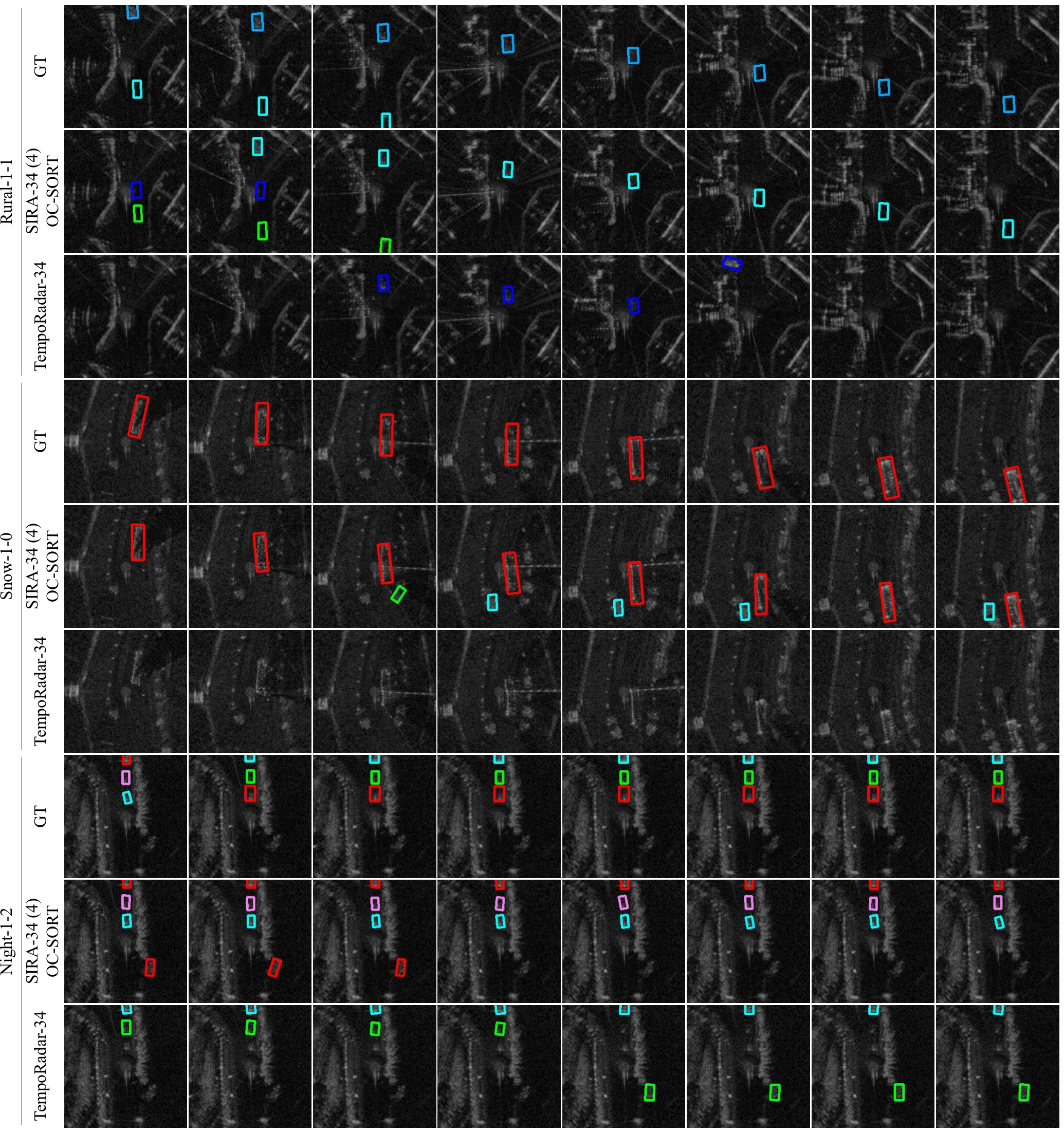}
    \caption{
        \textbf{Sampled tracking results} on three scenarios: Rural-1-1 (Top $3$ Rows), Snow-1-0  (Middle $3$ Rows) and Night-1-2 (Bottom $3$ Rows) on \textit{Radiate}. For each scenario, we include ground truth (GT), SIRA (SIRA-34 (4)) and TempoRadar (TempoRadar-34). The color of bounding boxes represents the object ID. The column represents consecutive radar frames. 
    }
    \label{fig:results_vis_tracking_3}
\end{figure*}

\section{Additional Ablation Study}
\label{sec:additional_ablation_study}
We present supplementary experimental results from our ablation studies.
Each experimental setting aligns with the conditions detailed in Section~\ref{sec:experiments}.

\paragraph{Detection Results at mAP@0.7:}
For Table~\ref{tab:results_detection} in Section~\ref{sec:result}, additional results for mAP@0.7 are shown in Table~\ref{tab:results_detection_full_evaluation}.
Compared to mAP@0.3 and mAP@0.5, all mAP@0.7 values are lower as expected when the IoU threshold increases. 
It is seen that SIRA provides consistently better detection performance than the baseline methods.

\paragraph{Detection Results With Full Size Radar Frames:}
We keep the original resolution with full size $1152\times 1152$ to make a fair comparison to the results from \cite{Sheeny2021_RADIATE}.
Regarding variations in image size, a marginal decline in detection performance is observed when dealing with larger scopes from Table \ref{tab:results_detection_comparison_non_crop}. 
However, empirical evidence has shown that the utilization of SIRA consistently leads to superior performance compared to the TemporRadar (TR).  

\paragraph{Tracking Results with Complete MOT Metrics:}
In Section~\ref{sec:result}, we showed tracking results in Table~\ref{tab:track_main} with selected metrics. Here, we show the tracking results with complete metrics including MOTP, FP, FN and ML~\cite{milan2016_mot16}.
The tracking results are shown in Table~\ref{tab:track_full_metrics} for comparison of the trackers.
ID switches vary based on \# of predicted BBox. With increased false negatives (FNs), both \# of BBoxes and ID switches reduce. Table~\ref{tab:track_full_metrics} suggests that TR generates more FNs and SIRA fewer FNs, thus higher ID switches. Nevertheless, we highlight the high IDF1 score in Table~\ref{tab:track_main} and Table~\ref{tab:track_full_metrics} of SIRA.

\paragraph{Number of Frames on Tracking:} 
According to Table~\ref{tab:track_frames_full_metrics}, considering longer time horizon contributes to the improvement in tracking performance in metrics such as  MOTA and IDF1. These results clarify the significance of extending to longer time horizon while maintaining computational scalability.
Fig.~\ref{fig:comparison_graph} illustrates the benefits of integrating more radar frames for the tracking performance over a range of methods. Compared with the single-frame baseline (ResNet-34 CenterTrack), SIRA with $T=6$ consecutive frames results in a margin of $+34.67$ MOTA. 

\paragraph{Effect of $\lambda$ in (\ref{eq:MCTrack}):}
Fig.~\ref{fig:comparison_lambda} illustrates the results obtained by varying $\lambda$ in (\ref{eq:MCTrack}) in the main paper, which corresponds to $C^{\text{angle}}$ when $\lambda=1$ and $C^{\text{tracklet}}$ when $\lambda=0$. Fig.~\ref{fig:comparison_lambda} appears to suggest that a combination of $C^{\text{angle}}$ and $C^{\text{tracklet}}$, i.e., $\lambda \in [0.3, 0.7]$,  consistently improves the tracking performance.

\paragraph{Performance of the Pseudo-Direction Estimation:}
We evaluated the pseudo-direction estimation performance in the terms of the smooth $L_1$ loss in (\ref{eq:DEst_lost_2}) over the test dataset. Fig.~\ref{fig:tracking_loss} shows the loss histogram for three time steps $\tau=T-1/T-2/T-3$ and it confirms that the majority of estimation errors are close to $0$, indicating a high accuracy.

\section{Visualization Results}
\label{sec:additional_visualization_results}

\paragraph{Detection with Cropped Radar Frames:}
Fig.~\ref{fig:results_vis_detection_1} visualizes the detection results of Table~\ref{tab:results_detection} in Section~\ref{sec:result} in adverse weather conditions: Night-1-4, Fog-6-0, and Rain-4-0, where green boxes represent the ground truth and red ones are the predictions.
In this case, with $T=4$ consecutive radar frames, SIRA allows for less FNs (unpaired green boxes) and less FPs (unpaired red boxes) in the BBox prediction than TempoRadar.

\paragraph{Detection with Full Size Radar Frames:}
Sampled detection results of Table~\ref{tab:results_detection_comparison_non_crop} are visualized in Fig.~\ref{fig:vis_imgs_non_crops}. 
SIRA demonstrates robust performance even when applied to full size radar frames. However, compared to the cropped frames, there is a slight decline in performance with full size radar frames. Upon closer investigation of this phenomenon, it is observed that object shapes in regions distant from the radar appear blurred due to lower angular resolution, leading to a slight increase in both FPs and FNs. Furthermore, this blurring increases the difficulty of predicting angles, resulting in a lower IoU. FP predictions are also attributed to ghost objects present in the radar signal, as pointed out by Li et al.~\cite{Li2022_TemporalRelations}. 

\paragraph{Tracking:}
Fig.~\ref{fig:results_vis_tracking_1} is in good weather, and Fig.~\ref{fig:results_vis_tracking_2} and Fig.~\ref{fig:results_vis_tracking_3} are in bad weather. 
From Fig.~\ref{fig:results_vis_tracking_1}, TempoRadar faces by numerous FNs and frequent ID switches. In contrast, SIRA, leveraging longer temporal information for consideration of spatio-temporal consistency, exhibits fewer FNs and a reduced ID switches. As a result, SIRA consistently achieves stable tracking. Moreover, Fig.~\ref{fig:results_vis_tracking_2} illustrates that SIRA can detect and track objects even in adverse weather conditions. Particularly in the Rain-4-0 environment, where vehicles exhibit nonlinear movement, the continuous tracking without interruptions underscores the effectiveness of MCTrack. 
However, in Fig.~\ref{fig:results_vis_tracking_3}, SIRA does exhibit a slight presence of FPs, likely influenced by reflections from multipath or ghost objects, due to tracking across consecutive frames. Addressing such false information poses an intriguing challenge. 


\clearpage

\section{KF-based Multiple Object Tracking for Radar Perception}
\label{sec:fundamentals_MOT}

\paragraph{Kalman Filter}
KF is a linear estimator for discretized dynamical systems in the time domain.
KF operates by utilizing state estimations from the previous time step and current measurements to predict the target state at the next time step. 
The filter maintains two key variables: the posterior state estimate represented as $\B{x}$, and the posterior estimate covariance matrix denoted as $\B{P}$. 

In the context of object tracking, the KF process is defined by several components, including the state transition model $\B{F}$, the observation model $\B{H}$, the process noise covariance $\B{Q}$, and the measurement noise covariance $\B{R}$. 
In each time step $t$, when presented with observations $\B{z}_t$, the KF operates through a sequence of predict and update stages.
\begin{align}
    \label{eq:kalman_filtering}
    & \text {predict}\left\{\begin{array}{l}
    \widehat{\B{x}}_{t \mid t-1} = \B{F}_t \widehat{\B{x}}_{t-1 \mid t-1} \\
    \B{P}_{t \mid t-1} = \B{F}_t \B{P}_{t-1 \mid t-1} \B{F}_t^{\top}+\B{Q}_t,
    \end{array}\right. \\
    & \text {update}\left\{\begin{array}{l}
    \B{K}_t = \B{P}_{t \mid t-1} \B{G}_t^{\top}\LS{\B{G}_t \B{P}_{t \mid t-1} \B{G}_t^{\top}+\B{R}_t}^{-1} \\
    \widehat{\B{x}}_{t \mid t} = \widehat{\B{x}}_{t \mid t-1}+\B{K}_t\LS{\B{z}_t-\B{G}_t \widehat{\B{x}}_{t \mid t-1}} \\
    \B{P}_{t \mid t} = \LS{\B{I}-\B{K}_t \B{G}_t} \B{P}_{t \mid t-1}
    \end{array} .\right.
\end{align}
The prediction stage involves calculating the state estimations for the subsequent time step $t$. In contrast, the update stage is focused on refining the posterior parameters within the KF when presented with measurfor thents of target states for time step $t$. 
In many scenarios, this measurement is derived from the observation model $\B{H}$ and is commonly referred to as an observation.

\paragraph{KF parameters}

In MOT, KF-based typically consists of five steps: Prediction, Association, Update, Deletion, and Initialization. 
The prediction and update phases are handled by KF.
In our setting for radar perception, the KF’s state $\B{x}_t$ and observation $\B{z}_t$ is defined as follows:
\begin{align}
    \B{x}_t &:=\LS{x_t, y_t, s_t, r_t, \vartheta_t, \dot{x}_t, \dot{y}_t, \dot{s}_t, \dot{\vartheta}_t}^{\top},\\
    \B{z}_t &:= \left(x_t, y_t, \widehat{w}_t, \widehat{h}_t, \widehat{\vartheta}_t, \widehat{c}_t \Mid \widehat{c}_t > \gamma \right)^{\top},    
\end{align}
where $\LS{x_t, y_t}$ is the two-dimensional coordinates of the object center in the image. $s = w\times h$ is the bounding box scale (area), $r$ is the bounding box aspect ratio and $\vartheta$ is object orientation, where $w$ and $h$ are the width and height of the object. 
The aspect ratio $r = \frac{w}{\text { float }(h+1 \mathrm{e}-6)}$ is assumed to be constant.
The other four variables, $\dot{x}$, $\dot{y}$, $\dot{s}$ and $\dot{\vartheta}$ are the corresponding time derivatives.
The detection confidence is $c$.
The observation model is
\begin{equation}
    \B{G}_t=\left[\begin{array}{lllllllll}
    1 & 0 & 0 & 0 & 0 & 0 & 0 & 0 & 0 \\
    0 & 1 & 0 & 0 & 0 & 0 & 0 & 0 & 0 \\
    0 & 0 & 1 & 0 & 0 & 0 & 0 & 0 & 0 \\
    0 & 0 & 0 & 1 & 0 & 0 & 0 & 0 & 0 \\
    0 & 0 & 0 & 0 & 1 & 0 & 0 & 0 & 0
    \end{array}\right]
\end{equation}
We note the process noise as in practice: $\B{Q}_t=\operatorname{diag}\LS{\sigma_{x}^2, \sigma_{y}^2, \sigma_s^2, \sigma_r^2, \sigma_{\vartheta}^2, \sigma_{\dot{u}}^2, \sigma_{\dot{v}}^2, \sigma_{\dot{s}}^2, \sigma_{\dot{\vartheta}}^2}$.
In the practice of SORT, we have to suppress the noise from velocity terms because it is too sensitive. 
We achieve it by setting a proper value for the process noise: 
\begin{equation}
    \B{Q}_t=\operatorname{diag}\LS{0.1, 5, 1^{-4}, 1^{-4}, 10, 0.01, 0.01, 1^{-4}, 0.1}.
\end{equation}
We note the linear transition model as:
\begin{equation}
    \B{F}_t=\left[\begin{array}{ccccccccc}
    1 & 0 & 0 & 0 & 0 & 1 & 0 & 0 & 0 \\
    0 & 1 & 0 & 0 & 0 & 0 & 1 & 0 & 0 \\
    0 & 0 & 1 & 0 & 0 & 0 & 0 & 1 & 0 \\
    0 & 0 & 0 & 1 & 0 & 0 & 0 & 0 & 0 \\
    0 & 0 & 0 & 0 & 1 & 0 & 0 & 0 & 1 \\
    0 & 0 & 0 & 0 & 0 & 1 & 0 & 0 & 0 \\
    0 & 0 & 0 & 0 & 0 & 0 & 1 & 0 & 0 \\
    0 & 0 & 0 & 0 & 0 & 0 & 0 & 1 & 0 \\
    0 & 0 & 0 & 0 & 0 & 0 & 0 & 0 & 1
    \end{array}\right],
\end{equation}
We set the measurement noise covariance as: 
\begin{equation}
    \B{R}_t=10\B{I}_5.
\end{equation}
We need to choose an initial value for $\B{P}_{t-1 \mid t-1}$, call it $\B{P}_{0 \mid 0}$.
If we were absolutely certain that our initial state estimate $\B{x}_0 = \B{0}$ was correct, we would let $\B{P}_{0 \mid 0}=\B{0}$. However, given the uncertainty in our initial estimate $\B{x}_0$, choosing $\B{P}_{0 \mid 0}=\B{0}$ would cause the filter to initially and always believe $\B{x}_t = \B{0}$. 
Assuming some uncertainty in the initial state, we set as follows: 
\begin{equation}
    \B{P}_{0 \mid 0}=\operatorname{diag}\LS{10, 10, 10, 10, 10, 10, 10, 10000, 10000},
\end{equation}
where $\dot{\vartheta}$ and $\dot{s}$ are set to a large value as the uncertainty is particularly high. On the other hand, we use the estimated pseudo-direction $\widehat{\B{d}}_{T|T-1}$ as the initial value for $\dot{u}, \dot{v}$. Therefore, we set small uncertainties for these.

\section{Fundamentals of FMCW for Automotive Radar}\label{sec:fundamentals_FMCW}

Radar technology offers a sensing solution that exhibits increased resilience to adverse weather conditions such as fog, rain, and snow. 
Typically, it generates low-resolution imagery, presenting significant challenges for tasks like object recognition and semantic segmentation. 
Contemporary automotive radar systems are primarily based on the Multiple Input Multiple Output (MIMO) technique, which employs multiple transmitters and receivers to determine the direction of arrival (DOA)~\cite{Texas2016_SRRRD}. 
Although this approach is cost-effective, existing configurations often suffer from limited azimuth resolution. 
For example, a commercial radar system with a 15$^{\circ}$ angular resolution produces a cross-range image with an approximate span of 10 meters at a distance of 20 meters. 
Consequently, radar imagery does not provide the level of detail necessary for effective object recognition and detailed scene mapping. 
On the contrary, the scanning radar employs a mobile antenna to measure azimuth at each point, leading to significantly improved azimuth resolution~\cite{Sheeny2021_RADIATE}.

\begin{figure}[t]
    \centering
    \includegraphics[width=0.485\textwidth]{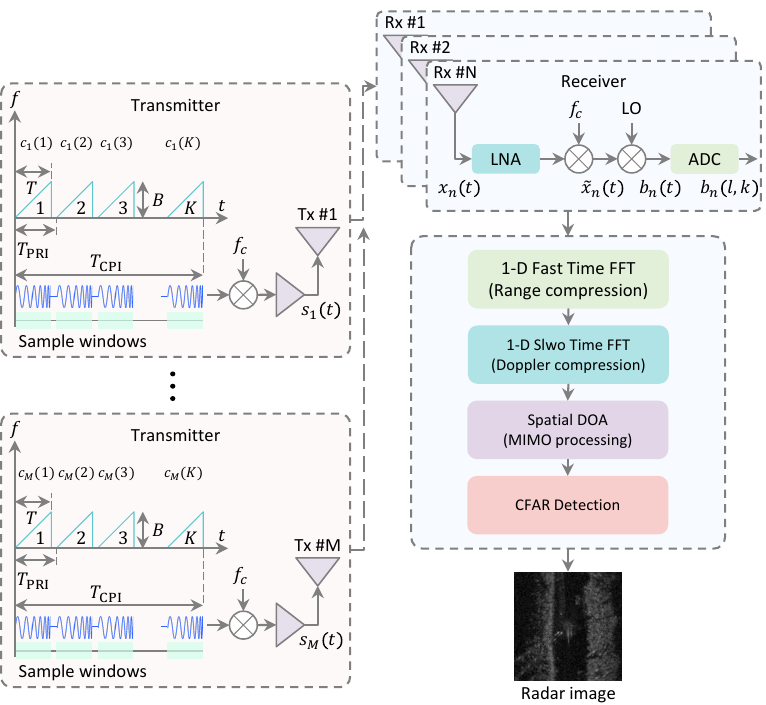}
    \caption{
        The slow-time FMCW automotive radar architecture from~\cite{Wang2020_GLRT}.
        On the left, a sequence of FMCW pulses with orthogonal slow-time (pulse) codes are sent from M transmitting antennas while, on the right, each of N receivers uses the same source FMCW waveform to sample the beat signal followed by range-doppler processing and slow-time waveform separation for spatial detection.
    }
    \label{fig:FMCW_architecture}
\end{figure}

\paragraph{Transmitter}
From~\cite{Wang2020_GLRT}, automotive radar predominantly employs a frequency-modulated continuous waveform (FMCW) for object detection, generating point clouds across multiple physical domains. 
As shown in Fig.~\ref{fig:FMCW_architecture}, this is achieved by transmitting a series of $K$ coded FMCW pulses from one of its $M$ Tx transmitting antennas, given by the expression of the radio frequency (RF) wave form on Tx antenna $m$:
\begin{align}
    \label{eq:transmitting_pulses}
    s_m(t) &= \sum_{k=0}^{K-1}{c_m\LS{k} s_p\LS{t-nT_{\text{PRI}}} e^{j2\pi f_c t}}, \\
    s_p\LS{t} &= \begin{cases}e^{j \pi \beta t^2} & 0 \leq t \leq T \\ 0 & \text { otherwise }\end{cases},
\end{align}
where $s_p\LS{t}$ is the baseband FMCW waveform (chirp pulse) with $\beta$ denoting the chirp rate and $T$ the pulse duration, and is repeated $K$ times.
$k$ is the index for pulse, and $c_m\LS{k}$ is the slow-time orthogonal code for the $k$-the pulse at the $m$-th Tx antenna, which satisfies the following:
\begin{equation}
    \sum_{k=0}^{K-1} c_i\LS{k} c_m\LS{k}= \begin{cases}K & \text { if } i=m \\ 0 & \text { otherwise }\end{cases}.
\end{equation}
$T_{\text{PRI}}$ is pulse repetition interval and $f_c$ is the carrier frequency, e.g., $f_c=79$ GHz.
The bandwidth of the FMCW waveform is $B = \beta T$.
The baseband waveform is repeated at each antenna before being multiplied by orthogonal codes $c_m\LS{k}$, for example, the Hadamard code.

\paragraph{Receiver}
An object at a range of $R_0$ with a radial velocity $v$ and a far-field spatial angle (i.e. azimuth, elevation, or both) induces amplitude attenuation and phase modulation to the received FMCW signal at each of $N$ Rx receiver RF chains (including the low noise amplifier (LNA), local oscillator (LO), and analog-to-digital converter (ADC)) of Fig.~\ref{fig:FMCW_architecture}.
The round-trip propagation delay from $m$-th Tx antenna to its $n$-th Rx antenna is
\begin{equation}
    \tau_{m n}\LS{t}=2 \frac{R_0+v t}{c}+m \frac{d_t \sin\LS{\theta_t}}{c}+n \frac{d_r \sin\LS{\theta_r}}{c},
\end{equation}
where $d_t$, $d_r$, $\theta_t$ and $\theta_r$ are the inter-element spacing and azimuthal angle for the transmitting and receiving antennas, respectively.
We assume co-located radars and the far-field approximation, i.e. $\theta_r = \theta_t = \theta$.
$c$ is the speed of propagation.
In the presence of an object at angle $\theta$, the $n$-th Rx receiver receives the signal of a sum of $M$ attenuated and delayed transmitting waveforms:
\begin{equation}
    x_{n}\LS{t}=\alpha \sum_{m=0}^{M-1} s_{m}\LS{t-\tau_{m n}} e^{j 2 \pi f_c\LS{t-\tau_{m n}}}.
\end{equation}
Subsequently, the baseband signal after LNA and carrier frequency down conversion is as follows:
\begin{align}
    \tilde{x}_n\LS{t} &= x_{n}\LS{t}e^{-j2\pi f_{c}t}\\
    &\approx \tilde{\alpha} \sum_{m=0}^{M-1} s_{m}\LS{t-\tau_0} e^{-j 2 \pi f_{c} \frac{2vt}{c}} e^{-j 2 \pi\LS{m d_t+n d_r} \frac{\sin \LS{\theta}}{\lambda}},
\end{align}
where $\tau_0 = \frac{2R_0}{c}$ is the time taken from the transmission to the reception, and $\lambda = \frac{c}{f_c}$ is wave length.
We assume that $s_{m}\LS{t-\tau_{mn}} = s_{m}\LS{t-\tau_0}$, and $\tilde{\alpha}$ absorbs constant phase factors.
By using LO, the signals at all receivers are mixed with the source chirp to generate the analog beat signal:
\begin{equation}
    b_n\LS{t}=\tilde{x}_n\LS{t} \sum_{k=0}^{K-1} s_{p}^*\LS{t-k T_{\mathrm{PRI}}},
\end{equation}
where $*$ denotes its conjugate.
This analog beat signal is then sampled at $t=k T_{\mathrm{PRI}}+l \Delta T$ with ADC sampling, where $\Delta T$ and $T_{\mathrm{PRI}}$ are the fast-time and slow-time sampling intervals, respectively, and digital beat signal is represented as follows:
\begin{equation}
    b_n\LS{l, k}=\tilde{\alpha} \sum_{m=0}^{M-1} c_m\LS{k} \underbrace{e^{-j 2 \pi f_r l}}_{\text{Range}} \underbrace{e^{-j 2 \pi f_d k}}_{\text{Doppler}} \underbrace{e^{-j 2 \pi\LS{f_s^t m + f_s^r n}}}_{\text{Virtual Spatial Array}},
\end{equation}
where $f_r=\LS{\beta \tau_0+2 f_c \frac{v}{c}} \Delta T$ is normalized range (fast-time) frequency, $f_d=2 f_c T_{\mathrm{PRI}} \frac{v}{c}$ is the normalized Doppler (slow-time) frequency, and $f_s^t$ and $f_s^r$ are the normalized spatial frequency at the transmitting and receiving antennas.
$f_s^t$ is usually different from $f_s^r$ due to different Tx/Rx spacings.
In other words, the beat signal $b_n\LS{l, k}$ at $n$-th receiver is the sum of the object responses originating from all transmitted waveforms, coded using $c_m\LS{k}$.
The beat signal at each of $N$ Rx receiver forms a matrix:
\begin{equation}
    \B{B}_n \!=\! \LL{
    \begin{array}{cccc}
        b_n\LS{1, 1} & b_n\LS{2, 1} &  \ldots & b_n\LS{L, 1} \\
        b_n\LS{1, 2} & b_n\LS{2, 1} & \ldots & b_n\LS{L, 2} \\
        \vdots & \vdots & \ddots & \vdots \\
        b_n\LS{1, K} & b_n\LS{2, 2} & \ldots & b_n\LS{L, K}
    \end{array}
    }.
\end{equation}

The induced modulation from the target is captured by the baseband signal processing block (including fast Fourier transforms (FFT) over range, Doppler, and spatial domains). 
All these processes lead to a multi-dimensional spectrum. With the constant false alarm rate (CFAR) detection step that compares the spectrum with an adaptive threshold, radar point clouds are generated in the range, Doppler, azimuth, and elevation domains~\cite{Li2008_MIMORadar,Bilik2019_AutonomousVehiclesSignalProcessingSolutions,Wang2020_GLRT,Li2022_TemporalRelations}. 
Considering the computing and cost constraints, automotive radar manufacturers may define the radar point clouds in a subset of the full four dimensions. For example, traditional automotive radar generates detection points in the range-Doppler domain, whereas some produce the points in the range-Doppler azimuth plane~\cite{Karthik2017_TexasInstruments}. 
In the \textit{Radiate} dataset~\cite{Sheeny2021_RADIATE} considered in this paper, the radar point cloud is defined in the range azimuth plane with a 360$^{\circ}$ field view. 
The resulting polar coordinate point cloud is further transformed into an ego-centric Cartesian coordinate system, then a standard voxelization can convert the point cloud into a radar frame as $I_t \in \R^{1 \times H\times W}$.

\end{document}